\documentclass[runningheads]{llncs}
\usepackage{graphicx}
\usepackage{subfigure} 

\usepackage{tikz}
\usepackage{comment}
\usepackage{amsmath,amssymb} 
\usepackage{color}
\usepackage{multirow}

\usepackage[accsupp]{axessibility}  

\usepackage[colorlinks,
            linkcolor=green,
            anchorcolor=green,
            citecolor=green]{hyperref}
            
\usepackage[width=122mm,left=12mm,paperwidth=146mm,height=193mm,top=12mm,paperheight=217mm]{geometry}

\begin{document}
\pagestyle{headings}
\mainmatter

\title{MVSTER: Epipolar Transformer for Efficient Multi-View Stereo} 

\titlerunning{MVSTER}
%
\author{Xiaofeng Wang\inst{1} \and
Zheng Zhu\inst{2} \and
Fangbo Qin\inst{1}\and
Yun Ye\inst{2}\and
Guan Huang\inst{2}\and
Xu Chi\inst{2}\and
Yijia He\inst{3}\and
Xingang Wang\inst{1}}
\authorrunning{X. Wang, Z. Zhu, F. Qin et al.}
%
\institute{ Institute of Automation, Chinese Academy of Sciences\\
\email{\{wangxiaofeng2020,qinfangbo2013,xingang.wang\}@ia.ac.cGn} \and
PhiGent Robotics \quad
\email{zhengzhu@ieee.org}\quad \email{\{yun.ye,guang.huan,xu.chi\}@phigent.ai}\and
Kwai Inc.\quad
\email{heyijia2016@gmail.com}}

\maketitle

\begin{abstract}
\sloppy
Learning-based Multi-View Stereo (MVS) methods warp source images into the reference camera frustum to form  3D volumes, which are fused as a cost volume to be regularized by subsequent networks. The fusing step plays a vital role in bridging 2D semantics and 3D spatial associations. However, previous methods utilize extra networks to learn 2D information as fusing cues, underusing 3D spatial correlations and bringing additional computation costs. 
Therefore, we present MVSTER, which leverages the proposed epipolar Transformer to learn both 2D semantics and 3D spatial associations efficiently. Specifically, the epipolar Transformer utilizes a detachable monocular depth estimator to enhance 2D semantics and uses cross-attention  to construct data-dependent 3D associations along epipolar line.  Additionally, MVSTER is built in a cascade structure, where entropy-regularized optimal transport is leveraged to propagate finer depth estimations in each stage. Extensive experiments show MVSTER achieves state-of-the-art reconstruction performance with significantly higher efficiency: Compared with MVSNet and CasMVSNet, our MVSTER achieves 34\% and 14\% relative improvements on the DTU benchmark, with 80\% and 51\% relative reductions in running time. 
MVSTER also ranks first on Tanks\&Temples-Advanced among all published works. Code is available at \url{https://github.com/JeffWang987/MVSTER}.
\end{abstract}

\section{Introduction}
\sloppy
Given multiple 2D RGB observations and camera parameters, Multi-View Stereo (MVS) aims to reconstruct the dense geometry of the scene.  MVS is a fundamental task in 3D computer vision, with applications ranging from autonomous navigation to virtual/augmented reality. Despite being extensively studied by traditional geometric methods \cite{DBLP:conf/iccv/GallianiLS15,DBLP:conf/cvpr/SchonbergerF16,DBLP:journals/mva/TolaSF12,DBLP:conf/cvpr/XuT19} for years, MVS is still challenged by unsatisfactory reconstructions under conditions of illumination changes, non-Lambertian surfaces and textureless areas \cite{DBLP:journals/tog/KnapitschPZK17,DBLP:conf/cvpr/SchopsSGSSPG17}.

Recent researches \cite{DBLP:conf/eccv/YaoLLFQ18,DBLP:conf/cvpr/0008LLSFQ19} have relieved the aforementioned problems via learning-based methods. Typically, they extract image features through 2D Convolutional  Neural Networks (CNN). Then, source features are warped into reference camera frustum to form source volumes, which are fused as a cost volume to produce depth estimations. Fusing source volumes is an essential step in the whole pipeline and many MVS approaches \cite{DBLP:conf/eccv/YaoLLFQ18,DBLP:conf/bmvc/ZhangYLLF20,wei2021aa,DBLP:conf/cvpr/WangGVSP21,DBLP:conf/eccv/YiWDZCWT20} put efforts into it. The core of the fusing step is to explore correlations between multi-view images. MVSNet \cite{DBLP:conf/eccv/YaoLLFQ18} follows the philosophy that various images contribute equally to the 3D cost volume, and utilizes variance operation to fuse different source volumes. However, such fusing method ignores various illumination and visibility conditions of different views. To alleviate this problem, \cite{DBLP:conf/cvpr/WangGVSP21,DBLP:journals/corr/abs-2111-14600,giang2021curvature} enrich 2D feature semnatics via Deformable Convolution Network (DCN) \cite{DBLP:conf/iccv/DaiQXLZHW17}, and \cite{DBLP:conf/eccv/YiWDZCWT20,DBLP:conf/bmvc/ZhangYLLF20} leverage extra networks to learn per-pixel weights as a guidance for fusing multi-view features. However, these methods introduce onerous network parameters and restrict efficiency. Besides, they only concentrate on 2D local similarities as a criteria for correlating multiple views, neglecting depth-wise 3D associations, which could lead to inconsistency in 3D space~\cite{DBLP:conf/cvpr/HeYFY20}.

Therefore, in this paper, we explore an efficient approach to model 3D spatial associations for fusing source volumes. Our intuition is to learn 3D relations from data itself, without introducing extra learning parameters. Recent success in attention mechanism prompts that Transformer \cite{DBLP:conf/nips/VaswaniSPUJGKP17} is appropriate for modeling 3D associations. The key advantage of Transformer is it leverages cross-attention to build data-dependent correlations, introducing minimal learnable parameters. Besides, compared with CNN, Transformer has expanded receptive field, which is more adept at constructing long-range 3D relations. Therefore, we propose the epipolar Transformer, which efficiently builds multi-view 3D correlations along the epipolar line. Specifically, we firstly leverage an auxiliary monocular depth estimator to enhance the 2D semantics of the $\textit{$\textit{query}$}$ feature. The auxiliary branch guides our network to learn depth-discriminative features, and it can be detached after training, which brings no extra computation cost. Subsequently, cross-attention is utilized to model 3D associations explicitly from features on epipolar lines, without introducing sophisticated networks. Additionally, we formulate the depth estimation as a depth-aware classification problem and solve it with entropy-regularized optimal transport \cite{DBLP:journals/ftml/PeyreC19}, which propagates finer depth maps in a cascade structure.

Owing to the epipolar Transformer, MVSTER obtains enhanced reconstruction results with fewer depth hypotheses. Compared with MVSNet \cite{DBLP:conf/eccv/YaoLLFQ18} and CasMVSNet \cite{DBLP:conf/cvpr/GuFZDTT20}, our method reduces 88\% and 73\% relative depth hypotheses, making 80\% and 51\% relative reduction in running time, yet obtaining 34\% and 14\% relative improvements on the DTU benchmark, respectively. Besides, our method ranks first among all published works on Tanks\&Temples-Advanced. The main technique contributions are four-fold as follows:

- We propose a novel end-to-end Transformer-based method for multi-view stereo, named MVSTER. It leverages the proposed epipolar Transformer to efficiently learn 3D associations along epipolar line.

- An auxiliary monocular depth estimator is utilized to guide  the  $\textit{query}$ feature to  learn  depth-discriminative information during training, which enhances feature semantics yet brings no efficiency compromises.

- We formulate depth estimation as a depth-aware classification problem and solve it with the entropy-regularized optimal transport, which produces finer depth estimations propagated in the cascade structure.

- Extensive experiments on DTU, Tanks\&Temples, BlendedMVS, and ETH3D show our method achieves superior performance with significantly higher efficiency than existing methods.

\section{Related Work}
\subsubsection{Learning-based MVS}
With the rapid progress of deep learning in 3D perception \cite{DBLP:conf/cvpr/QiSMG17,DBLP:conf/cvpr/ZhouT18,DBLP:conf/nips/QiYSG17,DBLP:conf/cvpr/ShiGJ0SWL20,DBLP:conf/cvpr/KeBASB17,DBLP:conf/eccv/MildenhallSTBRN20,DBLP:conf/iccv/DosovitskiyFIHH15,DBLP:conf/cvpr/LuoH21}, the MVS community is gradually dominated by learning-based methods \cite{DBLP:conf/eccv/YaoLLFQ18,DBLP:conf/cvpr/0008LLSFQ19,DBLP:conf/eccv/YanWYDZCWT20,DBLP:conf/cvpr/WangGVSP21,wei2021aa,ma2021epp,DBLP:conf/cvpr/GuFZDTT20,DBLP:conf/eccv/YiWDZCWT20}. They achieve better reconstruction results than traditional methods \cite{DBLP:conf/iccv/GallianiLS15,DBLP:conf/cvpr/SchonbergerF16,NeillDFCampbell2008UsingMH,YasutakaFurukawa2010AccurateDA,DBLP:journals/mva/TolaSF12}. Learning-based MVS approaches project source images into reference camera frustum to form multiple 3D volumes, which are fused through variance operation \cite{DBLP:conf/eccv/YaoLLFQ18,DBLP:conf/cvpr/0008LLSFQ19,DBLP:conf/eccv/YanWYDZCWT20,DBLP:conf/cvpr/GuFZDTT20,DBLP:conf/cvpr/YangMAL20,DBLP:conf/iccv/ChenHXS19}. Such a fusing method follows the philosophy that the feature volumes from various source images contribute equally \cite{DBLP:conf/eccv/YaoLLFQ18}, neglecting heterogeneous illumination and scene content variability \cite{DBLP:conf/eccv/YiWDZCWT20}. To remedy the aforementioned problem, PVA-MVSNet \cite{DBLP:conf/eccv/YiWDZCWT20} proposes a self-adaptive view aggregation module to learn the different significance in source volumes. Vis-MVSNet \cite{DBLP:conf/bmvc/ZhangYLLF20} computes pixel-visibility to represent matching quality, which serves as a volume fusing weight. AA-RMVSNet \cite{wei2021aa} leverages expensive DCNs \cite{DBLP:conf/iccv/DaiQXLZHW17} to enhance intra-view semantics, and it aggregates inter-view with pixel-wise weight. However, these methods use CNN-based module aggregating local features as fusing guidance, which lacks long-range 3D  associations and thus restricts their performance under 
challenging conditions. Besides, such aggregation modules bring extra computation cost burdening the network. In contrast, the proposed epipolar Transformer learns both 2D semantics and 3D spatial relations from data itself, without bringing onerous network parameters.

\subsubsection{Efficient MVS}
To construct an efficient MVS pipeline, cascade-structured methods \cite{DBLP:conf/cvpr/GuFZDTT20,DBLP:conf/cvpr/ChengXZLLRS20,DBLP:conf/cvpr/YangMAL20,ma2021epp} are proposed. They address MVS problem in a coarse to fine manner, assuming decreasing depth hypotheses along reference camera frustum at each stage. PatchmatchNet \cite{DBLP:conf/cvpr/WangGVSP21} and IterMVS \cite{DBLP:journals/corr/abs-2112-05126} further decrease hypothesized depth number and discard expensive 3D CNN regularization in the cascade structure. However, they achieve high efficiency with significant performance compromises. Additionally, cascade methods have difficulty to recover from errors introduced at coarse resolutions \cite{DBLP:conf/cvpr/GuFZDTT20}. In this paper, cascade structure is leveraged to boost efficiency, and optimal transport is utilized to produce finer depth estimations at each stage of the cascade structure.

\subsubsection{Transformers in 3D Vision}
Transformers \cite{DBLP:conf/nips/VaswaniSPUJGKP17,DBLP:conf/acl/AbnarZ20,DBLP:conf/acl/TenneyDP19,DBLP:conf/naacl/DevlinCLT19,radford2018improving} find their initial applications in natural language processing and have drawn attention from computer vision community \cite{DBLP:conf/iclr/DosovitskiyB0WZ21,liu2021Swin,DBLP:conf/eccv/CarionMSUKZ20,DBLP:conf/cvpr/YangYFLG20,DBLP:conf/icml/ChenRC0JLS20,DBLP:conf/eccv/WangZGAYC20}. In tasks for 3D vision, PYVA \cite{DBLP:conf/cvpr/YangLLYMHP21} and NEAT \cite{Chitta2021ICCV} use cross-attention to build correlations between bird's eye view and front view. STTR \cite{li2021revisiting} formulates stereo depth estimation as a sequence-to-sequence correspondence problem that is optimized by self-attention and cross-attention. Recently, Transformer extends its application to MVS. LANet \cite{DBLP:conf/wacv/ZhangHW0021}, TransMVSNet \cite{DBLP:journals/corr/abs-2111-14600} and MVSTR   \cite{DBLP:journals/corr/abs-2112-00336} introduce an attention mechanism extracting dense features with global contexts, which expands the network receptive field. However, these methods densely correlate each pixel within 2D feature maps, which makes significant efficiency compromises. On the contrary,  our epipolar Transformer leverages geometric knowledge, restricting attention associations within the epipolar line, which significantly reduces dispensable feature correlations and makes our pipeline more efficient. Besides, MVSTER only leverages the essential cross-attention of Transformer \cite{DBLP:conf/nips/VaswaniSPUJGKP17}, without introducing sophisticated architecture (\textit{i.e.}, position encoding, Feedforward Neural Network (FNN) and  self-attention), which further boosts efficiency.

\subsubsection{Auxiliary Task Learning}
Auxiliary branch learning is demonstrated effective in multiple vision tasks \cite{DBLP:conf/cvpr/HeZH0Z20,DBLP:conf/cvpr/ZhaoZZZ19,DBLP:conf/nips/MordanTHC18}. In general, the auxiliary tasks are selected to be positively related to the main task, thus taking effect during training. In addition, the branch can be discarded after training, bringing no burden during inference. ManyDepth \cite{DBLP:conf/cvpr/WatsonAPBF21} is a self-supervised monocular depth estimator, utilizing MVS cost volume as an auxiliary branch, which enhances estimation reliability. This inspires us that MVS assists monocular depth estimation, and vice versa. Therefore, an auxiliary monocular depth estimation branch is leveraged in MVSTER to learn depth-discriminative features.

\section{Method}
\begin{figure}
\centering
\includegraphics[height=4.0cm]{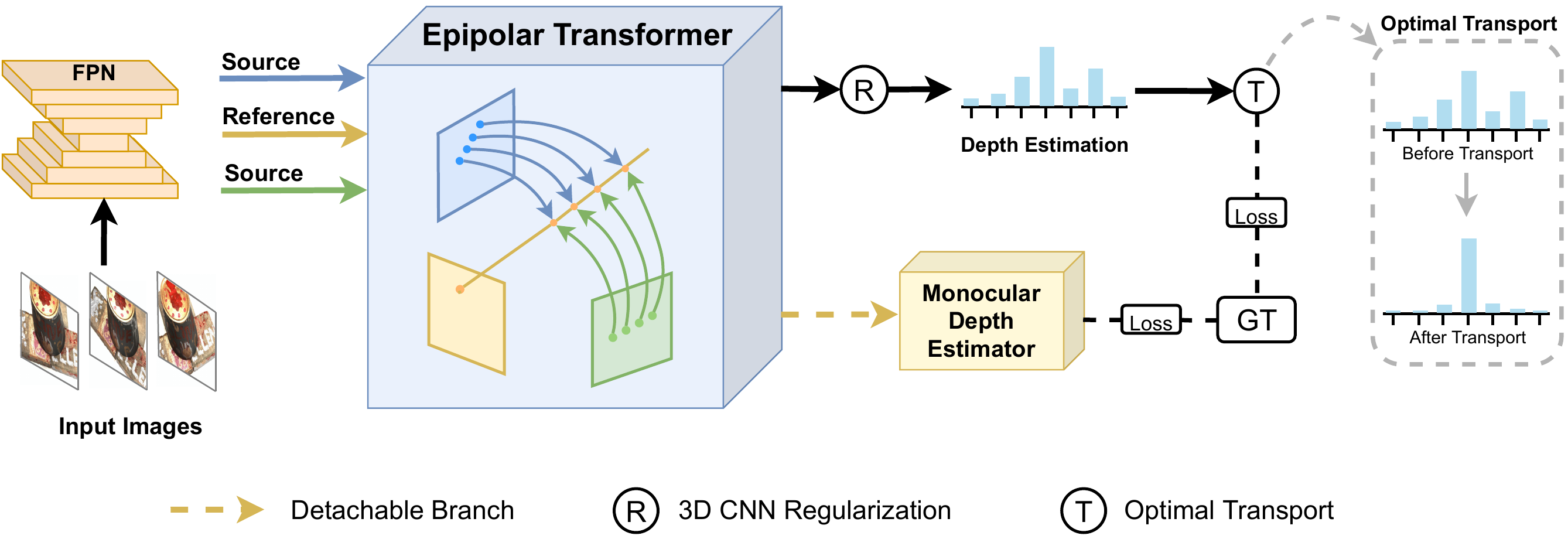}
\caption{MVSTER architecture. MVSTER firstly extracts features via FPN, then the multi-view features are aggregated by the epipolar Transformer, where the auxiliary branch makes monocular depth estimation to enhance context. Subsequently, the aggregated feature volume is regularized by 3D CNNs, producing depth estimations. Finally, optimal transport is utilized to optimize the predicted depth.}
\label{fig:mvster}
\end{figure}
In this section, we give a detailed description of MVSTER. The network architecture is illustrated in Fig.~\ref{fig:mvster}. Given a reference image and its corresponding source images, we firstly extract 2D multi-scale features using Feature Pyramid Network (FPN) \cite{DBLP:conf/cvpr/LinDGHHB17}. Source image features are then warped into reference camera frustum to construct source volumes via differentiable homography (Sec.~\ref{sec:3.1}). Subsequently, we leverage the epipolar Transformer to aggregate source volumes and produce the cost volume, which is regularized by lightweight 3D CNNs to make depth estimations (Sec.~\ref{sec:3.2}). Our pipeline is further built in a cascade structure, propagating depth map in a coarse to fine manner (Sec.~\ref{sec:3.3}). To reduce erroneous depth hypotheses during depth propagating, we formulate depth estimation as a depth-aware classification problem and optimize it with optimal transport. Finally, the network losses are given (Sec.~\ref{sec:3.4}).

\subsection{2D Encoder and 3D Homography}
\label{sec:3.1}

Given a reference image $\mathbf{I}_{i=0}\in \mathbb{R}^{H \times W \times 3}$ and its neighboring source images $\mathbf{I}_{i=1,...,N-1} \in \mathbb{R}^{H \times W \times 3}$, the first step is to extract the multi-scale 2D features of these inputs. A FPN-like network is applied, where the images are downscaled $M$ times to build deep features $\mathbf{F}_{i=0,...,N-1}^{k=0,...,M-1} \in \mathbb{R}^{H_k \times W_k \times C_k}$. The scale $k=0$ denotes the original size of images. The subsequent formulations can be generalized to a specific scale $k$, so $k$ is omitted for simplicity.

Following previous learning-based methods \cite{DBLP:conf/eccv/YaoLLFQ18,DBLP:conf/cvpr/0008LLSFQ19,DBLP:conf/cvpr/WangGVSP21,DBLP:journals/corr/abs-2111-14600}, we utilize plane sweep stereo \cite{RobertTCollins1996ASA} that establishes multiple front-to-parallel planes in the reference view. Specifically, equipped with camera intrinsic parameters $\left\{\mathbf{K}_i\right\}_{i=0}^{N-1}$ and transformations parameters $\left\{\left[\mathbf{R}_{0, i} \mid \mathbf{t}_{0, i}\right]\right\}_{i=1}^{N-1}$ from source views to reference view, source features can be warped into the reference camera frustum:

\begin{equation}
\label{eq:homo}
    \mathbf{p}_{s_i,j}=\mathbf{K}_{i} \cdot\left(\mathbf{R}_{0, i} \cdot\left({\mathbf{K}_{0}}^{-1} \cdot \mathbf{p}_r \cdot d_{j}\right)+\mathbf{t}_{0, i}\right) ,
\end{equation}
where $d_j$ denotes $j$-th hypothesized depth of pixel $\mathbf{p}_r$ in the reference feature, and $\mathbf{p}_{s_i,j}$ is the corresponding pixel in the $i$-th source features. After the warping operation, $N-1$ source volumes $\left\{\mathbf{V}_i\right\}_{i=1}^{N-1}\in\mathbb{R}^{H \times W \times C \times D}$ are constructed, where $D$ is the total number of hypothesized depths.

\subsection{Epipolar Transformer}
\label{sec:3.2}
Next, we introduce the epipolar Transformer to aggregate source volumes from different views. The original attention function in Transformer \cite{DBLP:conf/nips/VaswaniSPUJGKP17} can be described as mapping a $\textit{query}$ and a set of $\textit{key}$-$\textit{value}$ pairs to an output. Similarly, in the proposed epipolar Transformer, the reference feature is leveraged as the user $\textit{query}$ to match source features ($\textit{keys}$) along the epipolar line, thus enhancing the corresponding depth  $\textit{value}$. Specifically, we enrich the reference $\textit{query}$ via an auxiliary task of monocular depth estimation. Subsequently, cross-attention computes associations between $\textit{query}$ and source volumes under epipolar constraint, generating attention guidance to aggregate the feature volumes from different views. The aggregated features are then regularized by lightweight 3D CNNs. In the following, we firstly give details about the $\textit{query}$ construction, then elaborate on the epipolar Transformer guided feature aggregation. Finally, the lightweight regularization strategy is given.

\subsubsection{Query Construction}
As aforementioned, we deem the reference feature as a $\textit{query}$ for the epipolar Transformer. However, features extracted by shallow 2D CNNs become less discriminative at non-Lambertian and low-texture regions. To remedy this problem, \cite{DBLP:conf/cvpr/WangGVSP21,DBLP:journals/corr/abs-2111-14600,wei2021aa,giang2021curvature} utilize expensive DCNs \cite{DBLP:conf/iccv/DaiQXLZHW17} or ASPP \cite{sinha2020deltas} to enrich features. In contrast, we propose a more efficient way to enhance our $\textit{query}$: building an auxiliary monocular depth estimation branch to regularize the $\textit{query}$ and learn depth-discriminative features. 

A common decoder \cite{DBLP:conf/iccv/GodardAFB19} used in the monocular depth estimation task is applied in our auxiliary branch. Given multi-scale reference features $\left\{\mathbf{F}_{0}^{k}\right\}_{k=0}^{M-1} $ that are extracted via FPN, we expand a low resolution feature map through interpolation, and concatenate it with the subsequent scale feature. The aggregated feature maps are fed into regression head \cite{DBLP:conf/iccv/GodardAFB19,ClmentGodard2017UnsupervisedMD} to make monocular depth estimations:
\begin{equation}
    \mathbf{M}_k = \mathbf{\Phi}(\left[\mathbf{I}(\mathbf{F}_0^k),\mathbf{F}_0^{k+1}\right]),
\end{equation}
where $\mathbf{\Phi}(\cdot)$ is monocular depth decoder, $\mathbf{I}(\cdot)$ is the interpolation function and $[\cdot,\cdot]$ denotes concatenation operation. Subsequently, the monocular depth estimation is repeated for queries with different scales. Notably, such auxiliary branch is only used in the training phase, guiding our network to learn depth-aware features.

\subsubsection{Epipolar Transformer Guided Aggregation} 
\begin{figure}[htbp]
\centering
\subfigure[Epipolar Transformer guided aggregation]{
\includegraphics[width=10cm]{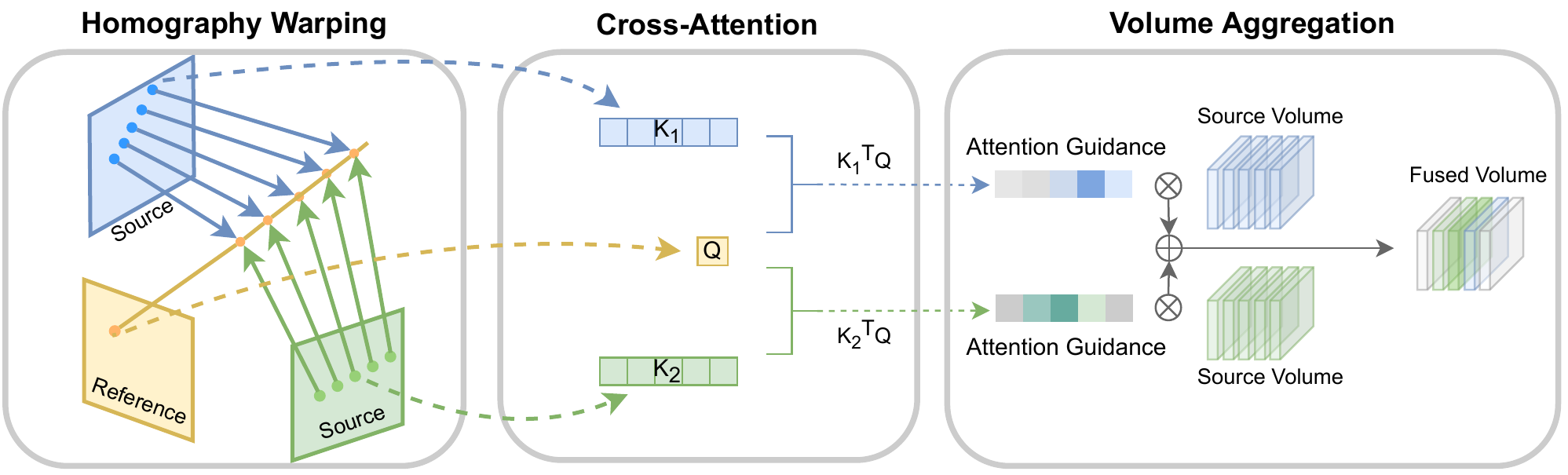}
\label{fig:et0}
}
\quad

\subfigure[Visualization of attention on the DTU dataset]{
\includegraphics[width=10cm]{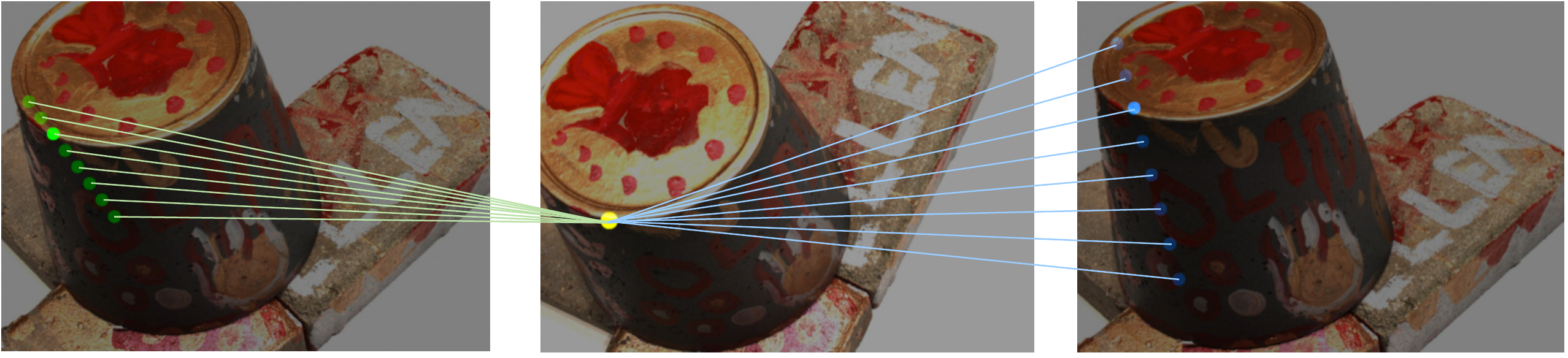}
\label{fig:et2}
}
\caption{(a) The epipolar Transformer aggregation. Homography warping is leveraged to restore depth-wise information of the reference feature, then cross-attention computes 3D associations between query and source volumes under epipolar constraint, generating attention guidance to aggregate the feature volumes from
different views. (b) Visualization of the cross-attention score on scan 1 of the DTU dataset, where the opacity of dots on the epipolar line represents the attention score.}
\label{fig:et}
\end{figure}

The aggregation pipeline is depicted in Fig.~\ref{fig:et0}, which aims at building 3D associations of the $\textit{query}$ feature. However, depth-wise 3D spatial information is not explicitly delivered by the 2D query feature map, so we firstly restore the depth information via homography warping. According to the warping operation in Equation~(\ref{eq:homo}), the hypothesized depth locations of $\textit{query}$ 
feature $\mathbf{p}_r$ are projected onto the source image epipolar line, resulting in the source volume features $\left\{\mathbf{p}_{s_i,j}\right\}_{j=0}^{D-1}$, which are regarded as the $\textit{keys}$ for the epipolar Transformer. Consequently, the $\textit{key}$ features along the epipolar line are leveraged to construct depth-wise 3D associations of the $\textit{query}$ feature, which is implemented with the cross-attention operation:

\begin{equation}
    \mathbf{w}_i = \text{softmax}({\frac{\mathbf{v_i}^\text{T} \mathbf{p}_r}{t_e\sqrt{C}}}),
    \label{eq:et0}
\end{equation}
where $\mathbf{v_i}\in \mathbb{R}^{C\times D}$ is calculated by stacking $\left\{\mathbf{p}_{s_i,j}\right\}_{j=0}^{D-1}$ along depth dimension, $t_e$ is the temperature parameter, and $\mathbf{w}_i$ is the attention correlating $\textit{query}$ and $\textit{keys}$. We visualize an example of real images in Fig.~\ref{fig:et2}, where the attention focuses on the most matched location on the epipolar line.

The calculated attention $\mathbf{w}_i$ between $\textit{query}$ and $\textit{keys}$ is utilized to aggregate $\textit{values}$. As for the Transformer $\textit{value}$ design, we follow \cite{DBLP:conf/cvpr/WangGVSP21,DBLP:conf/aaai/XuT20,DBLP:conf/bmvc/ZhangYLLF20} to use group-wise correlation, which measures the visual similarity between reference feature and source volumes in an efficient manner:

\begin{equation}
    \mathbf{s}_i^{g}=\frac{1}{G}\left\langle \mathbf{v}_i^g,\mathbf{p}_r^g\right\rangle,
    \label{eq:et1}
\end{equation}
where $g=0,...,G-1$, $\mathbf{v}_i^g\in \mathbb{R}^{\frac{C}{G}\times D}$ is the $g$-th group feature of $\mathbf{v}_i$, $\mathbf{p}_r^g\in \mathbb{R}^{ \frac{C}{G}\times 1}$ is the $g$-th group feature of $\hat{\mathbf{p}_r}$, and $\left\langle \cdot,\cdot\right\rangle$ is the inner product. $\left\{\mathbf{s}_i^g\right\}_{g=0}^{G-1}$ are then stacked along channel dimension to get $\mathbf{s}_i\in \mathbb{R}^{G \times D}$, which is the $\textit{value}$ for our Transformer. Finally, $\textit{values}$ are aggregated by epipolar attention score $\mathbf{w}_i$ to determine the final cost volume:
\begin{equation}
    \mathbf{c} = \frac{\sum_{i=1}^{N-1}\mathbf{w}_i\mathbf{s}_i}{\sum_{i=1}^{N-1}\mathbf{w}_i}.
    \label{eq:et2}
\end{equation}

 In summary, for the proposed epipolar Transformer, a detachable monocular depth estimation branch is firstly leveraged to enhance depth-discriminative 2D semantics, then the cross-attention between $\textit{query}$ and $\textit{keys}$ is utilized to construct depth-wise 3D associations. Finally, the combined 2D and 3D information serves as guidance for aggregating different views. As shown in Equation~(\ref{eq:et0})-(\ref{eq:et2}), the epipolar Transformer is designed as  an efficient aggregation module, where no learnable parameter is introduced, and the epipolar Transformer only learns data-dependent associations.

\subsubsection{Lightweight Regularization}
Due to non-Lambertian surfaces or object occlusions, the raw cost volume is noise-contaminated \cite{DBLP:conf/eccv/YaoLLFQ18}. To smoothen the final depth map, 3D CNNs are utilized to regularize the cost volume. Considering we have embedded 3D associations into the cost volume, depth-wise feature encoding is omitted in our 3D CNNs, which makes it more efficient. Specifically, we reduce convolution kernel size from $3\times3\times3$ to $3\times3\times1$, only aggregating cost volume along feature width and height. The regularized probability volume $\mathbf{P} \in \mathbb{R}^{H\times W \times D}$ is highly desirable in per-pixel depth confidence prediction, which is leveraged to make depth estimations in the cascade structure.
\subsection{Cascade Depth Map Propagation}
\label{sec:3.3}
Cascade structure is proven effective in stereo depth estimation \cite{DBLP:conf/cvpr/ShenDR21,DBLP:conf/iccv/DuggalWMHU19,DBLP:conf/cvpr/TankovichH0KFB21}, monocular reconstruction \cite{bozic2021transformerfusion} and MVS \cite{DBLP:conf/cvpr/GuFZDTT20,DBLP:conf/cvpr/ChengXZLLRS20,DBLP:conf/cvpr/WangGVSP21},  which brings efficiency and enhanced performance. Following \cite{DBLP:conf/cvpr/WangGVSP21}, a four-stage searching pipeline is set for MVSTER, where the resolutions of inputs for the four stages are $H\times W\times 64$, $\frac{H}{2}\times \frac{W}{2}\times 32$, $\frac{H}{4}\times \frac{W}{4}\times 16$ and $\frac{H}{8}\times \frac{W}{8}\times 8$ respectively. Following \cite{DBLP:conf/aaai/XuT20,DBLP:conf/cvpr/WangGVSP21}, the inverse depth sampling is utilized to initialize depth hypotheses in the first stage, which is equivalent to equidistant sampling in pixel space. To propagate depth map in a coarse to fine manner, the depth hypotheses of each stage are centered at the previous stage's depth prediction, and $D_k$ hypotheses are uniformly generated within the hypothesized depth range. 

\subsection{Loss}
\label{sec:3.4}

\begin{figure}
\centering
\includegraphics[height=3.5cm]{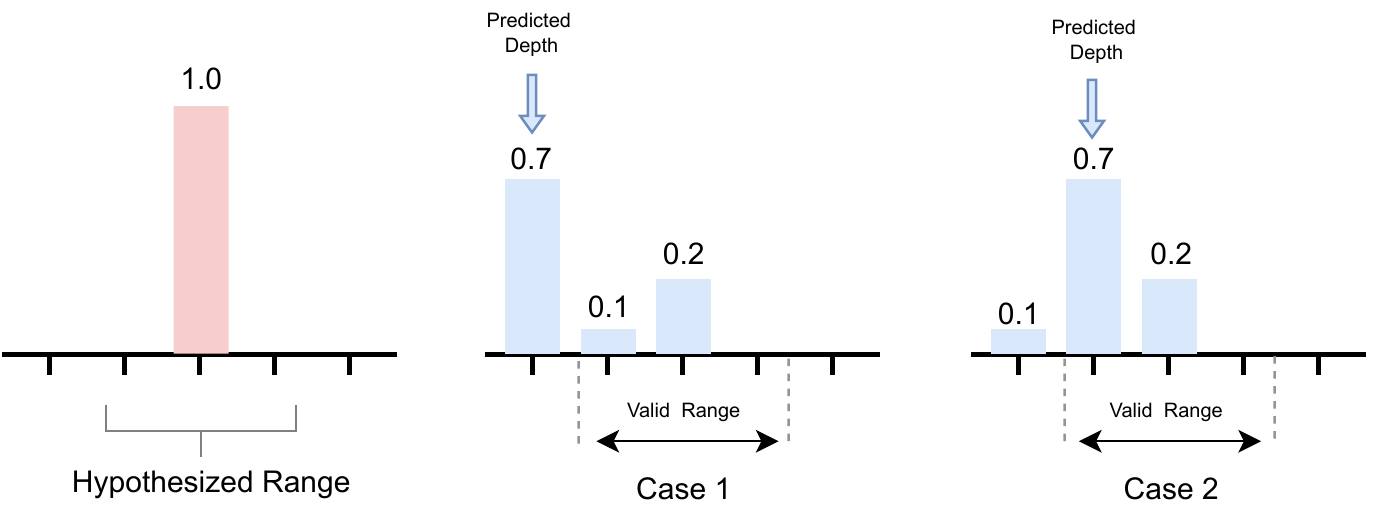}
\caption{Example illustrating that cross-entropy loss is not aware of the relative distance between each hypothesized depth. The left-most subfigure is the ground truth. Case 1, Case 2 are two predicted depth distributions.}
\label{fig:loss}
\end{figure}

Although cascade structure benefits from coarse to fine pipeline, it has difficulty recovering from errors introduced at previous stages \cite{DBLP:conf/cvpr/GuFZDTT20}. To alleviate this problem, a straightforward way is to generate a finer depth map at each stage, especially avoiding predicting depth far away from the ground truth. However, previous methods \cite{DBLP:conf/cvpr/0008LLSFQ19,DBLP:journals/corr/abs-2111-14600,wei2021aa} simply regard depth estimation as a multi-class classification problem, which treats each hypothesized depth equally without considering the distance relationship between them. For example in Fig.~\ref{fig:loss}, given a ground truth depth probability distribution, the cross-entropy losses of case 1 and case 2 are the same. However, the depth prediction of case 1 is out of the valid range and can not be properly propagated to the next stage.

In this paper, the depth prediction is formulated as a depth-aware classification problem, which emphasizes the penalty of the predicted depth that is distant from the ground truth. Specifically, we measure the distance between the predicted distribution $\mathbf{P}_{i}\in \mathbb{R}^D$ and the ground truth distribution $\mathbf{P}_{\theta,i}\in \mathbb{R}^D$ with the off-the-shelf Wasserstein distance \cite{DBLP:journals/corr/ArjovskyCB17}: 
\begin{equation}
    d_w(\mathbf{P}_{i},\mathbf{P}_{\theta,i})=\inf _{\gamma \in \Pi(\mathbf{P}_{i},\mathbf{P}_{\theta,i})} \sum_{x, y}|x-y | \gamma(x, y),
\end{equation}
where $\inf$ stands for infimum, and $\Pi(\mathbf{P}_{i},\mathbf{P}_{\theta,i})$ is the set of all possible distributions whose marginal distributions are $\mathbf{P}_{i}$ and $\mathbf{P}_{\theta,i}$, which satisfies $\sum_{x} \gamma(x, y)=\mathbf{P}_{i}(y)$ and  $\sum_{y} \gamma(x, y)=\mathbf{P}_{\theta,i}(x)$. Such formulation is inspired by the  optimal transport problem \cite{DBLP:journals/ftml/PeyreC19} that calculates the minimum work transporting $\mathbf{P}_{i}$ to $\mathbf{P}_{\theta,i}$, which can be differentially solved via the sinkhorn algorithm \cite{DBLP:conf/nips/Cuturi13}.

In summary, the loss function consists of two parts: Wasserstein loss measuring the distance between predicted depth distribution and ground truth, and $L_1$ loss optimizing monocular depth estimation:
\begin{equation}
    Loss=\sum_{k=0}^{M-1}\sum_{i\in\mathbf{p}_\text{valid}}{d_w(\mathbf{P}_{i}^k,\mathbf{P}_{\theta,i}^k)+\lambda L_{1}(\textbf{M}_i^k,\mathbf{P}_{\theta,i}^k)},
\end{equation}
where $\mathbf{p}_\text{valid}$ refers to the set of valid ground truth pixels, and $\lambda$ is the loss weight. The total loss is calculated for $M$ stages.

\section{Experiments}
\subsection{Datasets}
MVSTER is evaluated on DTU \cite{DBLP:journals/ijcv/AanaesJVTD16}, Tanks\&Temples \cite{DBLP:journals/tog/KnapitschPZK17}, BlendedMVS \cite{DBLP:conf/cvpr/0008LLZRZFQ20} and ETH3D \cite{DBLP:conf/cvpr/SchopsSGSSPG17} to verify its effectiveness. Among the four datasets, DTU is an indoor dataset under laboratory conditions, which contains 124 scenes with 49 views and 7 illumination conditions. 
Following MVSNet \cite{DBLP:conf/eccv/YaoLLFQ18}, DTU is split into \texttt{training}, \texttt{validation} and \texttt{test} set. Tanks\&Temples is a public benchmark providing realistic video sequences, which is divided into the intermediate set and a more challenging advanced set. BlendedMVS is a large-scale synthetic dataset that contains 106 \texttt{training} scans and 7 \texttt{validation} scans. ETH3D benchmark introduces high-resolution images with strong view-point variations, which is split into \texttt{training} and \texttt{test} sets. As for the evaluation metrics, DTU, Tanks\&Temples, and ETH3D evaluate point cloud reconstructions using overall metrics \cite{DBLP:journals/ijcv/AanaesJVTD16} and $F_1$ score \cite{DBLP:journals/tog/KnapitschPZK17,DBLP:conf/cvpr/SchopsSGSSPG17}. BlendedMVS evaluates depth map estimations using depth-wise metric \cite{DBLP:conf/cvpr/0008LLZRZFQ20}: EPE stands for $L_1$ distance between predicted depth map and ground truth, $e_1$ and $e_3$ represent the proportion of pixels  with depth error larger than 1 and 3.

\subsection{Implementation Details}
\label{sec:id}
Following the common practice \cite{DBLP:journals/corr/abs-2201-01501,DBLP:journals/corr/abs-2112-05126,DBLP:journals/corr/abs-2111-14600},  MVSTER is firstly trained on DTU \texttt{training} set and evaluated on DTU \texttt{test} set, then it is finetuned on BlendedMVS before being tested on Tanks\&Temples and ETH3D benchmark. For DTU training, we use ground truth provided by MVSNet \cite{DBLP:conf/eccv/YaoLLFQ18}, whose depth range is sampled from 425$\rm{mm}$ to 935$\rm{mm}$. The input view selection and data pre-processing are the same as \cite{DBLP:conf/cvpr/WangGVSP21}. For BlendedMVS, we use the original image resolution and the number of input images is set as 7.

The hypothesized depth numbers $\left\{D_k\right\}_{k=0,...,3}$ for each stage are set as 8, 8, 4, 4. Following \cite{DBLP:journals/corr/abs-2201-01501,wei2021aa}, the hypothesized number of the 1st stage is doubled when MVSTER is tested on Tanks\&Temples and ETH3D. The group correlation $\left\{G_k\right\}_{k=0,...,3}$ are set as 8, 8, 4, 4. For inverse depth sampling, the inverse depth range $R_k$ satisfies $\frac{1}{R_k}=\frac{1}{D_{k-1}-1}\frac{1}{R_{k-1}}$. For the epipolar Transformer, the temperature parameter $t_e$ is set as 2. And the loss weight $\lambda$ is set as 0.0003 in the experiments. We train MVSTER for 10 epochs and optimize it with Adam \cite{DBLP:journals/corr/KingmaB14} ($\beta_1=0.9,\beta_2=0.999$). MVSTER is trained on four NVIDIA RTX 3090 GPUs with batch size 2 on each GPU. The learning rate is initially set as 0.001, which decays by a factor of 2 after 6, 8 and 9 epochs. 

For point cloud reconstruction, we follow previous methods \cite{DBLP:conf/cvpr/0008LLSFQ19,DBLP:conf/eccv/YaoLLFQ18,DBLP:conf/cvpr/GuFZDTT20} to use both geometric and photometric constraints for depth filtering. We set the view consistency number and the photometric  probability threshold as 4 and 0.5, respectively.
The final depth fusion steps also follow previous methods \cite{DBLP:conf/cvpr/0008LLSFQ19,DBLP:conf/eccv/YaoLLFQ18,DBLP:conf/cvpr/GuFZDTT20}.

\subsection{Benchmark Performance}
\subsubsection{Evaluation on DTU} 
\begin{figure}
\centering
\includegraphics[height=3cm]{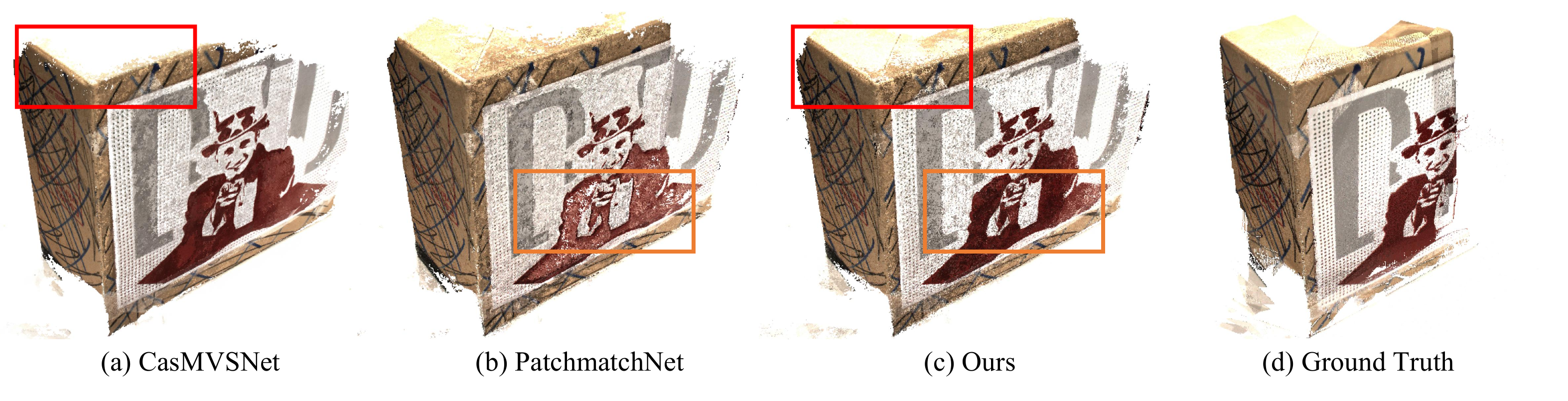}
\caption{Reconstruction results on DTU (scan 13). Our method delivers more accurate boundaries in the red bounding box than CasMVSNet \cite{DBLP:conf/cvpr/GuFZDTT20}, and preserves more detials of textureless area in the orange bounding box than PatchmatchNet \cite{DBLP:conf/cvpr/WangGVSP21}. }
\label{fig:visdtu}
\end{figure}

We compare MVSTER with traditional methods \cite{DBLP:conf/iccv/GallianiLS15,DBLP:conf/cvpr/SchonbergerF16,DBLP:journals/mva/TolaSF12}, published learning-based methods \cite{DBLP:conf/eccv/YaoLLFQ18,DBLP:conf/cvpr/0008LLSFQ19,DBLP:conf/cvpr/YuG20,DBLP:conf/eccv/YiWDZCWT20,DBLP:conf/cvpr/GuFZDTT20,DBLP:conf/cvpr/ChengXZLLRS20,DBLP:conf/cvpr/WangGVSP21,wei2021aa} and approaches from recent technical reports \cite{DBLP:journals/corr/abs-2111-14600,DBLP:journals/corr/abs-2112-00336,DBLP:journals/corr/abs-2201-01501,DBLP:journals/corr/abs-2112-05126}. The input images are set as different resolutions (MVSTER*: $1600\times1200$ and MVSTER: $864 \times 1152$) to compare with previous methods, and the number of views is set as 5. The quantitative results are shown in Table~\ref{table:dtu}, where MVSTER* achieves a state-of-the-art overall score and completeness score among all the competitors. Significantly, the inference time of MVSTER* is 0.17s, which is faster than the previous fastest method \cite{DBLP:conf/cvpr/WangGVSP21}. Additionally, MVSTER with lower resolution ($864 \times 1152$) still outperforms all published works, and it runs at 0.09s per image with 2764 MB GPU memory consumption, which sets a new state of the art for efficient learning-based MVS. Qualitative comparisons are shown in Fig.~\ref{fig:visdtu}, where MVSTER reconstructions provide denser results with finer details. Especially, compared with CasMVSNet \cite{DBLP:conf/cvpr/GuFZDTT20} and PatchmatchNet \cite{DBLP:conf/cvpr/WangGVSP21}, MVSTER recovers more details at object boundaries and textureless areas.

\setlength{\tabcolsep}{4pt}
\begin{table}
\scriptsize
\begin{center}
\caption{Quantitative results of different methods on the DTU \texttt{evaluation} set. Methods with * denote their input resolution is $1600\times1200$. The last four methods with gray font come from technical reports.}
\label{table:dtu}
\begin{tabular}{lcccc}
\hline\noalign{\smallskip}
Method & Acc.$\downarrow$ & Comp.$\downarrow$ & Overall$\downarrow$ & Runtime (s)$\downarrow$ \\
\noalign{\smallskip}
\hline
\noalign{\smallskip}
Gipuma \cite{DBLP:conf/iccv/GallianiLS15} & \textbf{0.283} & 0.873 & 0.578 & - \\
COLMAP \cite{DBLP:conf/cvpr/SchonbergerF16} & 0.400 & 0.664 & 0.532 & - \\
Tola \cite{DBLP:journals/mva/TolaSF12} & 0.342 & 1.190 & 0.766 & - \\
\noalign{\smallskip}
\hline
\noalign{\smallskip}
MVSNet \cite{DBLP:conf/eccv/YaoLLFQ18} & 0.396 & 0.527 & 0.462 & 0.85 \\
R-MVSNet \cite{DBLP:conf/cvpr/0008LLSFQ19} & 0.383 & 0.452 & 0.417 & 0.89 \\
Fast-MVSNet \cite{DBLP:conf/cvpr/YuG20} & 0.336 & 0.403 & 0.370 & 0.37 \\
CVP-MVSNet* \cite{DBLP:conf/eccv/YiWDZCWT20} & 0.296 & 0.406 & 0.351 & 1.12 \\
CasMVSNet \cite{DBLP:conf/cvpr/GuFZDTT20} & 0.325 & 0.385 & 0.355 & 0.35 \\
UCS-Net* \cite{DBLP:conf/cvpr/ChengXZLLRS20} & 0.338 & 0.349 & 0.344 & 0.32 \\
PatchmatchNet* \cite{DBLP:conf/cvpr/WangGVSP21} & 0.427 & 0.277 & 0.352 & 0.18 \\
AA-RMVSNet \cite{wei2021aa} & 0.376 & 0.339 & 0.357 & - \\
\noalign{\smallskip}
\hline
\noalign{\smallskip}
MVSTER & 0.350 & 0.276 & 0.313 & \textbf{0.09} \\
MVSTER* & 0.340 & \textbf{0.266}  & \textbf{0.303} & 0.17\\
\noalign{\smallskip}
\hline
\noalign{\smallskip}
\textcolor{gray}{TransMVSNet \cite{DBLP:journals/corr/abs-2111-14600}} & \textcolor{gray}{0.321} & \textcolor{gray}{0.289} & \textcolor{gray}{0.305} & \textcolor{gray}{0.99} \\
\textcolor{gray}{MVSTR \cite{DBLP:journals/corr/abs-2112-00336}} & \textcolor{gray}{0.356} & \textcolor{gray}{0.295} & \textcolor{gray}{0.326} & \textcolor{gray}{0.81} \\
\textcolor{gray}{UniMVSNet \cite{DBLP:journals/corr/abs-2201-01501}} & \textcolor{gray}{0.352} & \textcolor{gray}{0.278} & \textcolor{gray}{0.315} & \textcolor{gray}{-} \\
\textcolor{gray}{IterMVS* \cite{DBLP:journals/corr/abs-2112-05126}} & \textcolor{gray}{0.373} & \textcolor{gray}{0.354} & \textcolor{gray}{0.363} & \textcolor{gray}{0.18} \\

\hline
\end{tabular}
\end{center}
\end{table}
\setlength{\tabcolsep}{1.4pt}

\setlength{\tabcolsep}{2pt}
\begin{table}
\scriptsize
\begin{center}
\caption{Quantitative results on Tanks\&Temples-advanced. The evaluation metric is the mean F-score and the last four methods with gray font come from technical reports.}
\label{table:tankadv}
\begin{tabular}{lccccccc}
\hline\noalign{\smallskip}
Method & \textbf{Mean F-score}&Aud.&Bal.&Cou.&Mus.&Pal.&Tem.\\
\noalign{\smallskip}
\hline
\noalign{\smallskip}
COLMAP \cite{DBLP:conf/cvpr/SchonbergerF16}  & 27.24 &  16.02 &  25.23  & 34.70 &  41.51  & 18.05 &  27.94\\
ACMH \cite{DBLP:conf/cvpr/XuT19} & 34.02 & 23.41 & 32.91 & $\mathbf{41.17}$ & 48.13 & 23.87 & 34.60\\
\noalign{\smallskip}
\hline
\noalign{\smallskip}
R-MVSNet \cite{DBLP:conf/cvpr/0008LLSFQ19} & 29.55 &  19.49 &  31.45  & 29.99  & 42.31 &  22.94  & 31.10\\
CasMVSNet  \cite{DBLP:conf/cvpr/GuFZDTT20} & 31.12 &  19.81  & 38.46 &  29.10 &  43.87 &  27.36  & 28.11\\
PatchmatchNet \cite{DBLP:conf/cvpr/WangGVSP21}& 32.31  & 23.69  & 37.73 &  30.04  & 41.80 &  28.31 &  32.29\\
EPP-MVSNet \cite{ma2021epp} & 35.72 &  21.28  & 39.74  & 35.34 &  49.21  & 30.00  & 38.75\\
AA R-MVSNet \cite{wei2021aa} & 33.53 &  20.96  & 40.15 &  32.05 &  46.01 &  29.28  & 32.71\\
\noalign{\smallskip}
\hline
\noalign{\smallskip}

MVSTER &$\mathbf{37.53}$  &  $\mathbf{26.68}$  & $\mathbf{42.14}$  & 35.65 &  $\mathbf{49.37}$  & $\mathbf{32.16}$  & $\mathbf{39.19}$\\
\noalign{\smallskip}
\hline
\noalign{\smallskip}

\textcolor{gray}{TransMVSNet \cite{DBLP:journals/corr/abs-2111-14600}} & \textcolor{gray}{37.00} & \textcolor{gray}{24.84} & \textcolor{gray}{$\mathbf{44.69}$} & \textcolor{gray}{34.77}& \textcolor{gray}{46.49} & \textcolor{gray}{$\mathbf{34.69}$} & \textcolor{gray}{36.62} \\
\textcolor{gray}{MVSTR \cite{DBLP:journals/corr/abs-2112-00336}} & \textcolor{gray}{32.85} & \textcolor{gray}{22.83} & \textcolor{gray}{39.04 } & \textcolor{gray}{33.87}& \textcolor{gray}{45.46 } & \textcolor{gray}{27.95} & \textcolor{gray}{27.97} \\
\textcolor{gray}{UniMVSNet \cite{DBLP:journals/corr/abs-2201-01501}} & \textcolor{gray}{$\mathbf{38.96}$} & \textcolor{gray}{$\mathbf{28.33}$} & \textcolor{gray}{44.36} & \textcolor{gray}{39.74}& \textcolor{gray}{$\mathbf{52.89}$} & \textcolor{gray}{33.80} & \textcolor{gray}{ 34.63} \\
\textcolor{gray}{IterMVS \cite{DBLP:journals/corr/abs-2112-05126}} & \textcolor{gray}{34.17} & \textcolor{gray}{25.90} & \textcolor{gray}{38.41} & \textcolor{gray}{31.16}& \textcolor{gray}{44.83} & \textcolor{gray}{29.59} & \textcolor{gray}{35.15} \\

\hline
\end{tabular}
\end{center}
\end{table}
\setlength{\tabcolsep}{1.4pt}

\subsubsection{Evaluation on Tanks\&Temples} 
MVSTER is tested on Tanks\&Temples to demonstrate the generalization ability. We use the original image resolution and set the number of views as 7. The depth range, camera parameters, and view selection strategies are aligned with PatchmatchNet \cite{DBLP:conf/cvpr/WangGVSP21}. And we follow the dynamic consistency checking method in depth filtering \cite{DBLP:conf/eccv/YanWYDZCWT20}. We compare MVSTER with traditional methods \cite{DBLP:conf/cvpr/SchonbergerF16,DBLP:conf/cvpr/XuT19}, published learning-based methods \cite{DBLP:conf/cvpr/0008LLSFQ19,DBLP:conf/cvpr/GuFZDTT20,DBLP:conf/cvpr/WangGVSP21,ma2021epp,wei2021aa}, and approaches from recent technique reports \cite{DBLP:journals/corr/abs-2111-14600,DBLP:journals/corr/abs-2112-00336,DBLP:journals/corr/abs-2201-01501,DBLP:journals/corr/abs-2112-05126}. Advanced set quantitative results are shown in Table~\ref{table:tankadv}, where MVSTER achieves the highest mean F-score among all published works, and the inference time per image is 0.26s. Although our performance is 1.4\% lower than the recent UniMVSNet \cite{DBLP:journals/corr/abs-2201-01501}, the inference speed of MVSTER is 3$\times$ faster than UniMVSNet\footnote{The inference time of UniMVSNet \cite{DBLP:journals/corr/abs-2201-01501} is not reported, but its baseline is CasMVSNet \cite{DBLP:conf/cvpr/GuFZDTT20}, whose inference speed is 3$\times$ slower than MVSTER.}. For Tanks\&Temples-Intermediate, MVSTER achieves a 60.92 mean F-score, which is 7.8\% better than the previous most efficient method \cite{DBLP:conf/cvpr/WangGVSP21}. Overall, MVSTER shows strong generalization ability with great efficiency.

\subsubsection{Evaluation on ETH3D} 
 For evaluation on the ETD3D dataset, the input images are resized to $1920\times 1280$ and the number of inputs is set as 7. The depth range, camera parameters, and view selection strategies are aligned with PatchmatchNet \cite{DBLP:conf/cvpr/WangGVSP21}. We compare MVSTER with traditional methods \cite{DBLP:conf/cvpr/SchonbergerF16,YasutakaFurukawa2010AccurateDA,DBLP:conf/cvpr/SchonbergerF16,DBLP:conf/cvpr/XuT19}, published learning-based methods~\cite{DBLP:conf/cvpr/WangGVSP21,lee2021patchmatchrl} and approaches from technique reports \cite{DBLP:journals/corr/abs-2007-07714,DBLP:journals/corr/abs-2112-05126}. The running time per image is 0.30s and the quantitative results are shown in Table~\ref{table:eth3d}. On the \texttt{training} set, MVSTER achieves better $F_1$-score than the most competitive traditional method ACMH \cite {DBLP:conf/cvpr/XuT19} and the recent IterMVS \cite{DBLP:journals/corr/abs-2112-05126}. On the \texttt{test} set, our method obtains 8.9\% improvement over the previous most efficient method \cite{DBLP:conf/cvpr/WangGVSP21}, which is comparable to the recent IterMVS \cite{DBLP:journals/corr/abs-2112-05126}. This demonstrates MVSTER can be well generalized to high-resolution images.

\setlength{\tabcolsep}{4pt}
\begin{table}
\scriptsize
\begin{center}
\caption{Quantitative results on the ETH3D benchmark, which is split into a \texttt{training} set and a \texttt{test} set. The last two methods with gray font come from technical reports.}
\label{table:eth3d}
\begin{tabular}{l|ccc|ccc}
\hline \multirow{2}{*}{\text { Methods }} & \multicolumn{3}{c|}{\text { Training set }} & \multicolumn{3}{c}{\text { Test set }} \\
\cline { 2 - 7 } & \text { Acc. }  & \text { Comp. } & $F_{1}$ \text {-score }  & \text { Acc. }  & \text { Comp. } & $F_{1}$ \text {-score } \\
\hline
 Gipuma \cite{DBLP:conf/cvpr/SchonbergerF16}  & 84.44 & 34.91 & 36.38 & 86.47 & 24.91 & 45.18 \\
  PMVS  \cite{YasutakaFurukawa2010AccurateDA} & 90.23 & 32.08 & 46.06 & 90.08 & 31.84& 44.16 \\
 COLMAP  \cite{DBLP:conf/cvpr/SchonbergerF16} & \textbf{91.85} & 55.13 & 67.66 & 91.97 & 62.98 & 73.01 \\
 ACMH \cite{DBLP:conf/cvpr/XuT19} & 88.94 & 61.59 & 70.71 & \textbf{89.34} & 68.62 & 75.89 \\
 
\hline

 PatchmatchNet \cite{DBLP:conf/cvpr/WangGVSP21}& 64.81 & 65.43 & 64.21 & 69.71 & 77.46 & 73.12 \\
 PatchMatch-RL \cite{lee2021patchmatchrl}& 76.05 & 62.22 & 67.78 & 74.48 & 72.89 & 72.38\\
\hline
MVSTER & 76.92 &  \textbf{68.08}  & \textbf{72.06}  & 77.09 & \textbf{82.47}   &  \textbf{79.01} \\

\hline
 \textcolor{gray}{PVSNet \cite{DBLP:journals/corr/abs-2007-07714}} & \textcolor{gray}{67.84} & \textcolor{gray}{69.66} & \textcolor{gray}{67.48} & \textcolor{gray}{66.41} & \textcolor{gray}{80.05} & \textcolor{gray}{72.08} \\
 \textcolor{gray}{IterMVS \cite{DBLP:journals/corr/abs-2112-05126}}  & \textcolor{gray}{79.79} & \textcolor{gray}{66.08} & \textcolor{gray}{71.69} & \textcolor{gray}{84.73} & \textcolor{gray}{76.49} & \textcolor{gray}{\textbf{80.06}}\\
 \hline
\end{tabular}
\end{center}
\end{table}
\setlength{\tabcolsep}{1.4pt}

\vspace{-1cm}
\subsection{Ablation Study}
The ablation study is conducted to analyze the effectiveness of each component, which is measured with DTU's point cloud reconstruction metric \cite{DBLP:journals/ijcv/AanaesJVTD16} and BlendedMVS's depth estimation metric \cite{DBLP:conf/cvpr/0008LLZRZFQ20}. Unless specified, the image resolutions for DTU and BlendedMVS are $864\times1152$ and $576\times 768$.

\subsubsection{Epipolar Transformer (ET)}
Existing methods for aggregating different views in learning-based MVS can be categorized as two types: ($\romannumeral1$) variance fusing \cite{DBLP:conf/eccv/YaoLLFQ18,DBLP:conf/cvpr/GuFZDTT20,DBLP:conf/cvpr/ChengXZLLRS20,DBLP:conf/cvpr/YangMAL20,DBLP:conf/cvpr/0008LLSFQ19}, ($\romannumeral2$)  CNN-based fusing \cite{wei2021aa,DBLP:conf/eccv/YiWDZCWT20,DBLP:conf/bmvc/ZhangYLLF20,DBLP:conf/cvpr/WangGVSP21}. In this experiment, the aforementioned two methods are compared with the ET module under three conditions\footnote{Three hypothesized depth number: ($\romannumeral1$) $D:192$ used by one-stage methods \cite{DBLP:conf/eccv/YaoLLFQ18,DBLP:conf/cvpr/0008LLSFQ19,wei2021aa}, ($\romannumeral2$) $D:48, 32, 8$ used by three-stage methods \cite{DBLP:conf/cvpr/GuFZDTT20,DBLP:journals/corr/abs-2111-14600}, and  ($\romannumeral3$) $D:8, 8, 4, 4$ used by MVSTER. All of these conditions follow implementation details described in Sec.~\ref{sec:id}.}. The quantitative results are concluded in Table~\ref{table:eta}. We observe that reducing hypothesized depth number can significantly decrease inference time. Compared with the hypothesized number used by MVSNet \cite{DBLP:conf/eccv/YaoLLFQ18} and CasMVSNet \cite{DBLP:conf/cvpr/GuFZDTT20}, our method relatively reduces $70\%$ and $ 53\%$ running time. However, the variance fusing strategy shows restricted improvement in the third condition with fewest hypothesized number. CNN-based fusing alleviates the problem by enhancing local visual similarity, but it relatively brings 45\%, 46\%, 89\%  computation cost in three cases. In contrast, the ET module shows consistent performance improvement under different hypothesized cases, which demonstrates that 3D spatial associations are beneficial for aggregating multi-view features. Significantly, the ET module learns data-dependent fusing guidance, introducing minimal network parameters and bringing no extra computation cost.
\vspace{-0.3cm}
\setlength{\tabcolsep}{2.2pt}
\begin{table}
\scriptsize
\begin{center}
\caption{Quantitative results of different fusing methods under conditions with different hypothesized depth numbers.}
\label{table:eta}
\begin{tabular}{lccccccc}
\hline\noalign{\smallskip}
 Method &Hypo. Num. & Overall$\downarrow$ & EPE $\downarrow$& $e_1$$\downarrow$&$e_3$$\downarrow$&Runtime (s)$\downarrow$&Param (M)$\downarrow$\\

\noalign{\smallskip}
\hline
\noalign{\smallskip}
Variance Fusion & 192& 0.460 &  1.62  & 19.34 &  9.84 & 0.40  &  0.34  \\
CNN Fusion  & 192 & 0.442  & 1.58  & \textbf{17.89} &   9.47 &0.58 & 0.35  \\
ET (Ours)  & 192 & \textbf{0.435} & \textbf{1.54}   &  17.93 & \textbf{9.32} & \textbf{0.40}   & \textbf{0.34}  \\
\noalign{\smallskip}
\hline
\noalign{\smallskip}
Variance Fusion &48,32,8 &0.335 &  1.28  & 14.82 &  7.55 & 0.28  &  0.93  \\
CNN Fusion &48,32,8  & 0.327  & \textbf{1.07}  & 14.33 &   7.03 & 0.41 & 0.94  \\
ET (Ours)   &48,32,8  & \textbf{0.323} &  1.09  &  \textbf{14.17} & \textbf{6.89} & \textbf{0.28}   & \textbf{0.93}  \\
\noalign{\smallskip}
\hline
\noalign{\smallskip}
Variance Fusion  &8,8,4,4   & 0.334 &  1.39  & 15.32 &  7.92 & 0.09   & 0.98  \\
CNN Fusion  &8,8,4,4 & 0.320  & 1.33  & \textbf{14.80} &   7.32 &0.17 & 1.01  \\
ET (Ours)   &8,8,4,4   & \textbf{0.313} & \textbf{1.31}   &  14.98 & \textbf{7.27} & \textbf{0.09}   & \textbf{0.98}  \\
\hline
\end{tabular}
\end{center}
\end{table}
\setlength{\tabcolsep}{1.4pt}

\vspace{-1.2cm}
\subsubsection{Monocular Depth Estimator (MDE)} 
\begin{figure}
\setlength{\abovecaptionskip}{1pt}
\centering
\includegraphics[height=2.3cm]{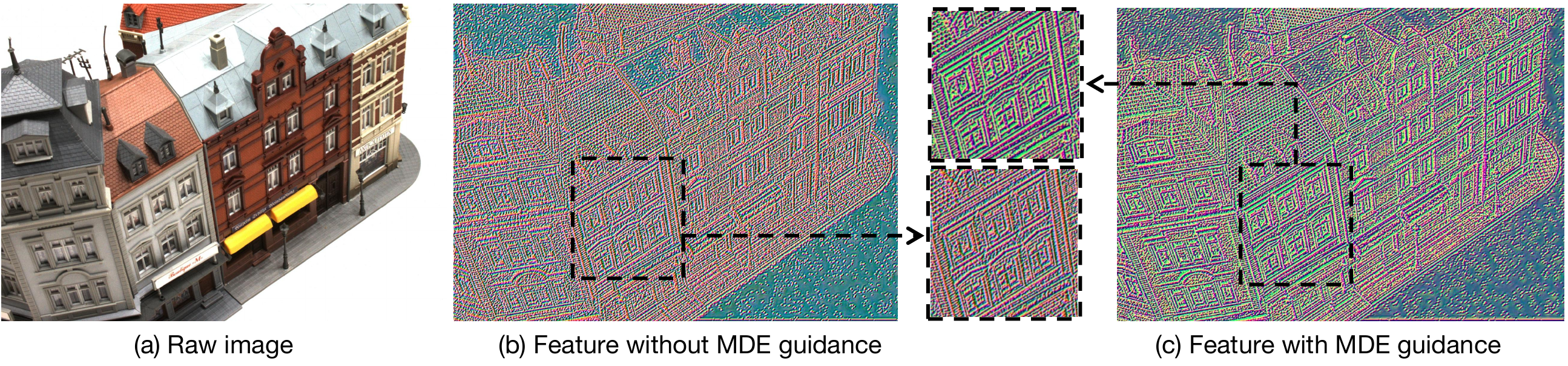}
\caption{An example shows that MDE guides the FPN feature to deliver more details at object boundaries. For better visualization, we leverage PCA to reduce the number of feature channels to 3 and color the channels with RGB.}
\label{fig:vismono}
\end{figure}

In this experiment, the proposed MDE module is compared with DCN used by \cite{DBLP:conf/cvpr/WangGVSP21,DBLP:journals/corr/abs-2111-14600,wei2021aa} and ASPP used by \cite{giang2021curvature}. The results are shown in Table~\ref{table:mdl}. We observe that the ASPP restricts the reconstruction performance, and DCN enhances reconstruction results with reduced depth error $e_1$. However, DCN brings high computation costs and introduces onerous learning parameters. In contrast, the MDE module shows comparable performance improvement with DCN but introduces no extra computation burden. We also provide an example in Fig.~\ref{fig:vismono}, where the MDE module enhances features details at object boundaries, which could reduce ambiguity for depth estimations within boundary areas.

\subsubsection{Optimal Transport in Depth Propagation (OT)}

Learning-based MVS methods usually use $L_1$ loss to regress the depth map \cite{DBLP:conf/eccv/YaoLLFQ18,DBLP:conf/cvpr/GuFZDTT20} or use cross-entropy loss to classify depth \cite{DBLP:conf/cvpr/0008LLSFQ19,DBLP:journals/corr/abs-2111-14600}. In this experiment, the two losses are compared with the Wasserstein loss computed by OT. Apart from the aforementioned evaluation metrics, we introduce $S_3$~EPE and $S_4$~EPE, which stand for EPE of stage 3 and stage 4 depth estimations on DTU. As shown in Table~\ref{table:ot}, OT improves point cloud reconstruction performance and reduces depth error. Especially, it greatly reduces depth error of the last two stages, which demonstrates OT is effective in propagating finer depth maps to later stages.
\vspace{-0.3cm}
\setlength{\tabcolsep}{4pt}
\begin{table}
\scriptsize
\begin{center}
\caption{Comparison of our MDE module with DCN and ASPP.}
\label{table:mdl}
\begin{tabular}{lcccccc}
\hline\noalign{\smallskip}
Method & Overall$\downarrow$ & EPE $\downarrow$& $e_1$$\downarrow$&$e_3$$\downarrow$&Runtime (s)$\downarrow$&Param (M)$\downarrow$\\

\noalign{\smallskip}
\hline
\noalign{\smallskip}
Raw feature & 0.317 & 1.35 &  15.00  &  7.37 &  0.09&  0.98  \\
DCN  & \textbf{0.313} &  1.33  & \textbf{14.82} &  7.55 & 0.23  &  1.51  \\
ASPP & 0.327  & 1.34  & 15.30 &   7.53 &0.15 & 1.16  \\
MDE (Ours)  & \textbf{0.313} & \textbf{1.31}   &  14.98 & \textbf{7.27} & \textbf{0.09}   & \textbf{0.98}  \\

\hline
\end{tabular}
\end{center}
\end{table}

\vspace{-1.5cm}

\vspace{-0.8cm}
\setlength{\tabcolsep}{4pt}
\begin{table}
\scriptsize
\begin{center}
\caption{Comparison of optimal transport with $L_1$ loss and cross-entropy loss.}
\label{table:ot}
\begin{tabular}{lcccccc}
\hline\noalign{\smallskip}
Method & Overall$\downarrow$ & EPE $\downarrow$& $e_1$ $\downarrow$ & $e_3$ $\downarrow$ &$S_3$ EPE $\downarrow$ & $S_4$ EPE$\downarrow$\\

\noalign{\smallskip}
\hline
\noalign{\smallskip}
$L_1$ Loss & 0.321 & 1.47 & 15.32  & 7.53 & 7.02& 6.32\\
CE Loss  & 0.314 & 1.34  &\textbf{14.96} & 7.40  & 7.12 &  6.64  \\
OT (Ours)  & \textbf{0.313} & \textbf{1.31}   &  14.98 & \textbf{7.27} &  \textbf{6.41}  & \textbf{5.90}  \\

\hline
\end{tabular}
\end{center}
\end{table}

\vspace{-1.2cm} 
\section{Conclusions}

In this paper, we present the epipolar Transformer for efficient MVS, termed as MVSTER. The proposed epipolar Transformer leverages both 2D semantics and 3D spatial associations to efficiently aggregate multi-view features. Specifically, MVSTER enriches 2D depth-discriminative semantics via an auxiliary monocular depth estimator. And the cross-attention on the epipolar line constructs 3D associations without learnable parameters. The combined 2D and 3D information serves as  guidance to aggregate different views. Moreover, we formulate depth estimation as a depth-aware classification problem, which produces finer depth estimations propagated in the cascade structure. Extensive experiments on DTU, Tanks\&Temple, BlendedMVS, and ETH3D show our method achieves stage-of-the-art performance with significantly higher efficiency. We hope that MVSTER can serve as an efficient baseline for learning-based MVS,  and further work may focus on simplifying 2D extractors and 3D CNNs.

\section{Additional Implementation Details}

\subsubsection{Network Architecture of Feature Extractor} We use a four-stage Feature Pyramid Network (FPN)~\cite{DBLP:conf/cvpr/LinDGHHB17} to extract image features, and the detailed parameters with layer descriptions are summarized in Table~\ref{table:fpn}. For Deformable Convolutional Networks (DCN)~\cite{DBLP:conf/iccv/DaiQXLZHW17} and Atrous Spatial Pyramid Pooling (ASPP)~\cite{DBLP:journals/corr/ChenPSA17} that are used in the ablation study of our main text, the network parameters are listed in Table~\ref{table:dcn}.

\setlength{\tabcolsep}{10pt}
\begin{table}
\scriptsize
\begin{center}
\caption{The detailed parameters of FPN, where S denotes stride, and if not specified with *, each convolution layer is followed by a Batch Normalization layer (BN) and a Rectified Linear Unit (ReLU).}
\label{table:fpn}
\begin{tabular}{lcc}
\hline\noalign{\smallskip}
Stage Description & Layer Description & Output Size \\
\noalign{\smallskip}
\hline
\noalign{\smallskip}
-& Input Images  & $H\times W\times 3$\\
\noalign{\smallskip}
\hline
\noalign{\smallskip}
  FPN Stage 1 & Conv2D, $3\times 3$, S1, 8  & $H\times W\times 8$\\
  \noalign{\smallskip}
  FPN Stage 1 & Conv2D, $3\times 3$, S1, 8  & $H\times W\times 8$\\
  \noalign{\smallskip}
  FPN Stage 1 Inner Layer* & Conv2D, $1\times 1$, S1, 64  & $H\times W\times 64$\\
  \noalign{\smallskip}
  FPN Stage 1 Output Layer* & Conv2D, $1\times 1$, S1, 8  & $H\times W\times 8$\\
\noalign{\smallskip}
\hline
\noalign{\smallskip}
  FPN Stage 2 & Conv2D, $5\times 5$, S2, 16  & $ H/2 \times W/2 \times 16$\\
  \noalign{\smallskip}
  FPN Stage 2 & Conv2D, $3\times 3$, S1, 16  & $ H/2 \times W/2 \times 16$\\
  \noalign{\smallskip}
  FPN Stage 2 & Conv2D, $3\times 3$, S1, 16  & $ H/2 \times W/2 \times 16$\\
  \noalign{\smallskip}
  FPN Stage 2 Inner Layer* & Conv2D, $1\times 1$, S1, 64  & $H/2\times W/2\times 64$\\
  \noalign{\smallskip}
  FPN Stage 2 Output Layer* & Conv2D, $1\times 1$, S1, 16  & $H/2\times W/2\times 16$\\
\noalign{\smallskip}
\hline
\noalign{\smallskip}
  FPN Stage 3 & Conv2D, $5\times 5$, S2, 32  & $H/4\times W/4\times 32$\\
  \noalign{\smallskip}
  FPN Stage 3 & Conv2D, $3\times 3$, S1, 32  & $H/4\times W/4\times 32$\\
  \noalign{\smallskip}
  FPN Stage 3 & Conv2D, $3\times 3$, S1, 32  & $H/4\times W/4\times 32$\\
  \noalign{\smallskip}
  FPN Stage 3 Inner Layer* & Conv2D, $1\times 1$, S1, 64  & $H/4\times W/4\times 64$\\
  \noalign{\smallskip}
  FPN Stage 3 Output Layer* & Conv2D, $1\times 1$, S1, 32  & $H/4\times W/4\times 32$\\
\noalign{\smallskip}
\hline
\noalign{\smallskip}
  FPN Stage 4 & Conv2D, $5\times 5$, S2, 64  & $H/8\times W/8\times 64$\\
  \noalign{\smallskip}
  FPN Stage 4 & Conv2D, $3\times 3$, S1, 64  & $H/8\times W/8\times 64$\\
  \noalign{\smallskip}
  FPN Stage 4 & Conv2D, $3\times 3$, S1, 64  & $H/8\times W/8\times 64$\\
  \noalign{\smallskip}
  FPN Stage 4 Inner Layer* & Conv2D, $1\times 1$, S1, 64  & $H/8\times W/8\times 64$\\
  \noalign{\smallskip}
  FPN Stage 4 Output Layer* & Conv2D, $1\times 1$, S1, 64  & $H/8\times W/8\times 64$\\
 \noalign{\smallskip}
\hline
\end{tabular}
\end{center}
\end{table}

\setlength{\tabcolsep}{6pt}
\begin{table}
\scriptsize
\begin{center}
\caption{The detailed parameters of DCN and ASPP, where S denotes stride, and D denotes dilation parameter for ASPP.  }
\label{table:dcn}
\begin{tabular}{lcc}
\hline\noalign{\smallskip}
Stage Description & Layer Description & Output Size \\
\noalign{\smallskip}
\hline
\noalign{\smallskip}
  DCN Stage 1 & DCN2D, $3\times 3$, S1, 8  & $H\times W\times 8$\\
  \noalign{\smallskip}
  DCN Stage 2 & DCN2D, $3\times 3$, S1, 16  & $H/2\times W/2\times  16$\\
  \noalign{\smallskip}
  DCN Stage 3 & DCN2D, $3\times 3$, S1, 32  & $H/4\times W/2\times  32$\\
  \noalign{\smallskip}
  DCN Stage 4  & DCN2D, $3\times 3$, S1, 64  & $H/8\times W/8\times  64$\\
  
\noalign{\smallskip}
\hline
\noalign{\smallskip}
  ASPP Stage 1 & Conv2D, $3\times 3$, S1, D\{1,6,12\}, 8  & $H\times W\times 8$\\
  \noalign{\smallskip}
  ASPP Stage 2 & Conv2D, $3\times 3$, S1, D\{1,6,12\}, 16  & $H/2\times W/2\times  16$\\
  \noalign{\smallskip}
  ASPP Stage 3 & Conv2D, $3\times 3$, S1, D\{1,6,12\}, 32  & $H/4\times W/2\times  32$\\
  \noalign{\smallskip}
  ASPP Stage 4  & Conv2D, $3\times 3$, S1, D\{1,6,12\}, 64  & $H/8\times W/8\times  64$\\
\noalign{\smallskip}
\hline
\end{tabular}
\end{center}
\end{table}

\subsubsection{Network Architecture of Light-Weight 3D CNN} An UNet~\cite{DBLP:conf/miccai/RonnebergerFB15} structured 3D CNN is applied for cost volume regularization at each stage, where the kernel size $3\times3\times3$ is partially replaced with $3\times3\times1$ in MVSTER for a more efficient pipeline. Apart from the input cost volume size, the network architectures are the same for each stage in the cascade structure, so we only report the detailed parameters of the 4th stage in Table~\ref{table:3D}.

\setlength{\tabcolsep}{5pt}
\begin{table}
\scriptsize
\begin{center}
\caption{The detailed parameters of 3D CNN, where S denotes stride, and if not specified with *, each convolution layer is followed by a Batch Normalization layer (BN) and a Rectified Linear Unit (ReLU).}
\label{table:3D}
\begin{tabular}{lcc}
\hline\noalign{\smallskip}
Stage Description & Layer Description & Output Size \\
\noalign{\smallskip}
\hline
\noalign{\smallskip}
-& Input Cost Volume & $H\times W\times4\times 8$\\
\noalign{\smallskip}
\hline
\noalign{\smallskip}
  UNet Stage 1 & Conv3D, $3\times 3\times 1$, S1, 8  & $H\times W\times4\times 8$\\
  \noalign{\smallskip}
  UNet Stage 1 & Conv3D, $3\times 3\times 1$, S2, 16  & $H/2\times W/2\times 4\times 16$\\
  \noalign{\smallskip}
  UNet Stage 1 & Conv3D, $3\times 3\times 3$, S1, 16  & $H/2\times W/2\times 4\times 16$\\
  \noalign{\smallskip}
  UNet Stage 1 Inner Layer & TransposeConv3D, $3\times 3\times 1$, S2, 8  & $H\times W\times 4\times 8$\\
  \noalign{\smallskip}
  UNet Stage 1 Output Layer* & TransposeConv3D, $3\times 3\times 3$, S1, 8  & $H\times W\times 4\times 8$\\
\noalign{\smallskip}
\hline
\noalign{\smallskip}
  UNet Stage 2 & Conv3D, $3\times 3\times1$, S2, 32  & $ H/4 \times W/4 \times4\times32$\\
  \noalign{\smallskip}
  UNet Stage 2 & Conv3D, $3\times 3\times3$, S1, 32  & $ H/4 \times W/4 \times4\times32$\\
  \noalign{\smallskip}
  UNet Stage 2 Inner Layer & TransposeConv3D, $3\times 3\times 1$, S2, 16  & $H/2\times W/2\times 4\times 16$\\
\noalign{\smallskip}
\hline
\noalign{\smallskip}
  UNet Stage 3 & Conv3D, $3\times 3\times1$, S2, 64  & $ H/8 \times W/8 \times4\times64$\\
  \noalign{\smallskip}
  UNet Stage 3 & Conv3D, $3\times 3\times3$, S1, 64  & $ H/8 \times W/8 \times4\times64$\\
  \noalign{\smallskip}
  UNet Stage 3 Inner Layer & TransposeConv3D, $3\times 3\times 1$, S2, 32  & $H/4\times W/4\times 4\times 32$\\
\noalign{\smallskip}
\hline
\end{tabular}
\end{center}
\end{table}

\section{Additional Ablation Study}
\subsubsection{Ablation Study on Hyperparameters} We conduct an ablation study on loss weight $\lambda$ and temperature parameter $t_e$. As shown in Table~\ref{table:lambda}, $\lambda=3\times10^{-4}$ is a proper loss weight for jointly optimizing monocular depth estimation and multi-view stereo. As shown in Table~\ref{table:te}, MVSTER produces a finer depth map when slowly increasing $t_e$, and the network shows best reconstruction performance on DTU when $t_e=2$.

\setlength{\tabcolsep}{6pt}
\begin{table}
\scriptsize
\begin{center}
\caption{Ablation study on loss weight $\lambda$.  }
\label{table:lambda}
\begin{tabular}{ccccccc}
\hline\noalign{\smallskip}
$\lambda$ & Acc.$\downarrow$ & Comp.$\downarrow$ & Overall$\downarrow$ & EPE$\downarrow$ &$e_1\downarrow$&$e_3\downarrow$  \\
\noalign{\smallskip}
\hline
\noalign{\smallskip}
  $1\times10^{-2}$ & 0.361 &0.312& 0.337  & 1.56 &16.47& 8.32 \\
  \noalign{\smallskip}
  $1\times10^{-3}$ & 0.354 &0.287& 0.321  & 1.33 &15.01& \textbf{7.26} \\
  \noalign{\smallskip}
  $3\times10^{-4}$ & \textbf{0.350} &0.276& \textbf{0.313}  & \textbf{1.31} &14.98& 7.27 \\
  \noalign{\smallskip}
  $1\times10^{-4}$ & 0.354 &\textbf{0.275}& 0.314  & 1.32 &\textbf{14.97}& 7.28 \\
\noalign{\smallskip}
\hline
\end{tabular}
\end{center}
\end{table}

\setlength{\tabcolsep}{6pt}
\begin{table}
\scriptsize
\begin{center}
\caption{Ablation study on temperature parameter $t_e$.  }
\label{table:te}
\begin{tabular}{ccccccc}
\hline\noalign{\smallskip}
$\lambda$ & Acc.$\downarrow$ & Comp.$\downarrow$ & Overall$\downarrow$ & EPE$\downarrow$ &$e_1\downarrow$&$e_3\downarrow$  \\
\noalign{\smallskip}
\hline
\noalign{\smallskip}
  0.5 & 0.354 &0.279& 0.317  & 1.33 &15.09& 7.52 \\
  \noalign{\smallskip}
  1.0 & 0.353 &0.279&  0.314  & 1.33 &15.03& 7.47 \\
  \noalign{\smallskip}
  2.0 & \textbf{0.350} &0.276& \textbf{0.313}  & \textbf{1.31} &14.98& 7.27 \\
  \noalign{\smallskip}
  3.0 & 0.353 &\textbf{0.274}& 0.314  & \textbf{1.31} &\textbf{14.89}& \textbf{7.08} \\
\noalign{\smallskip}
\hline
\end{tabular}
\end{center}
\end{table}

\section{Point Cloud Visualizations}
\begin{figure}
\setlength{\abovecaptionskip}{1pt}
\centering
\includegraphics[height=18cm]{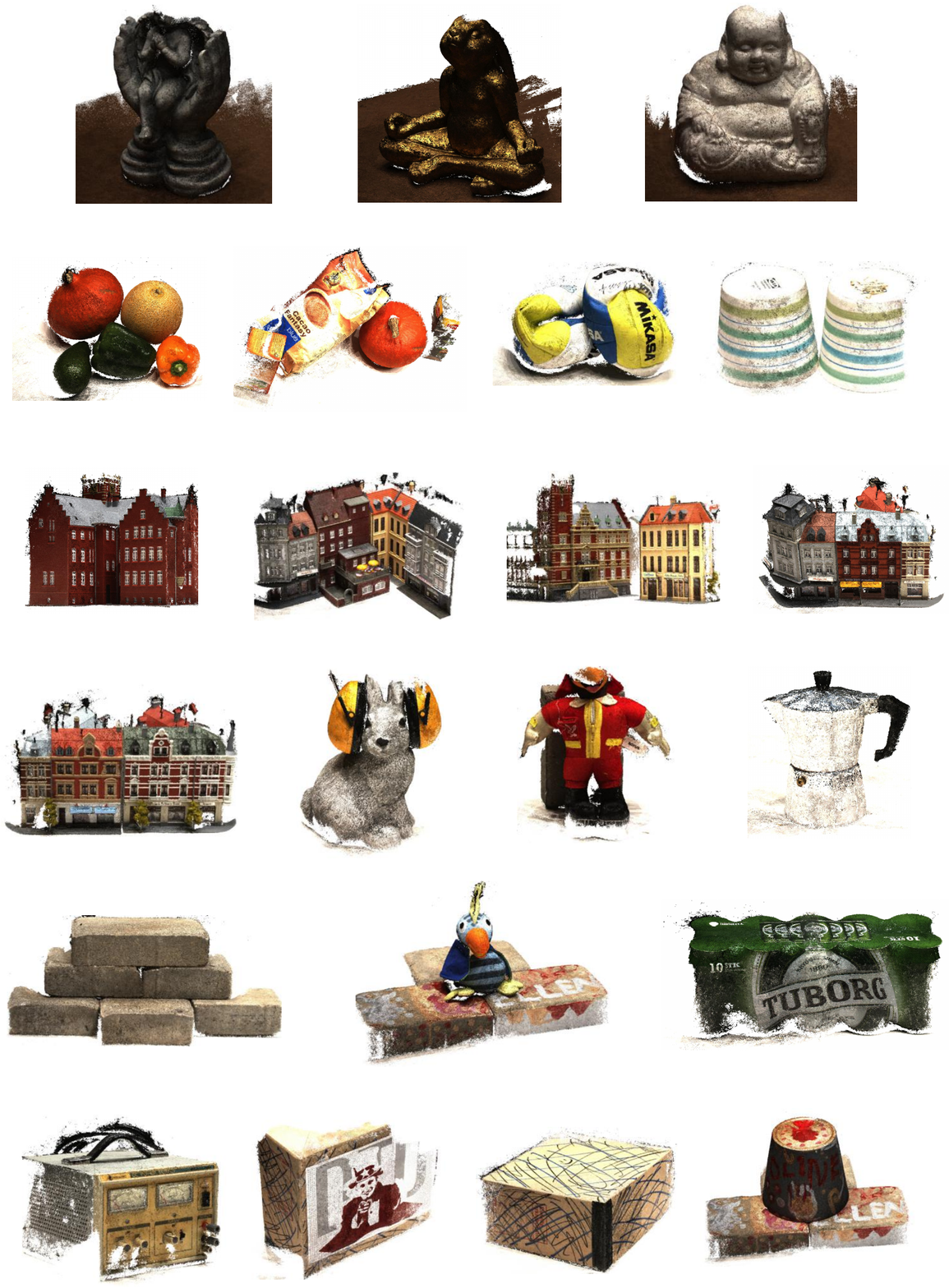}
\caption{Point clouds on DTU~\cite{DBLP:journals/ijcv/AanaesJVTD16} reconstructed by MVSTER.}
\label{fig:dtu}
\end{figure}

\begin{figure}
\setlength{\abovecaptionskip}{1pt}
\centering
\includegraphics[height=15cm]{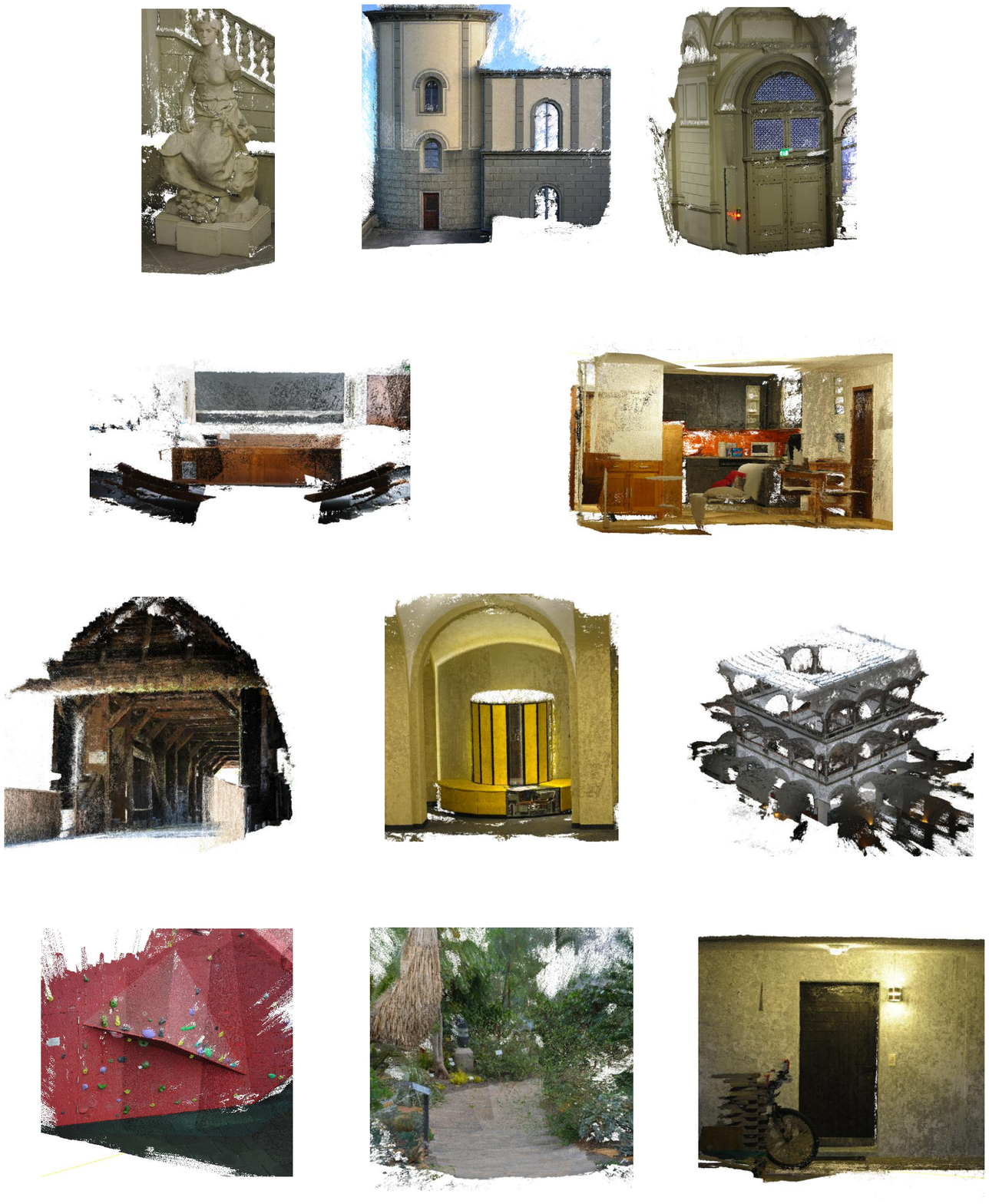}
\caption{Point clouds on ETH3D~\cite{DBLP:conf/cvpr/SchopsSGSSPG17} reconstructed by MVSTER.}
\label{fig:eth}
\end{figure}

\begin{figure}
\setlength{\abovecaptionskip}{1pt}
\centering
\includegraphics[height=16cm]{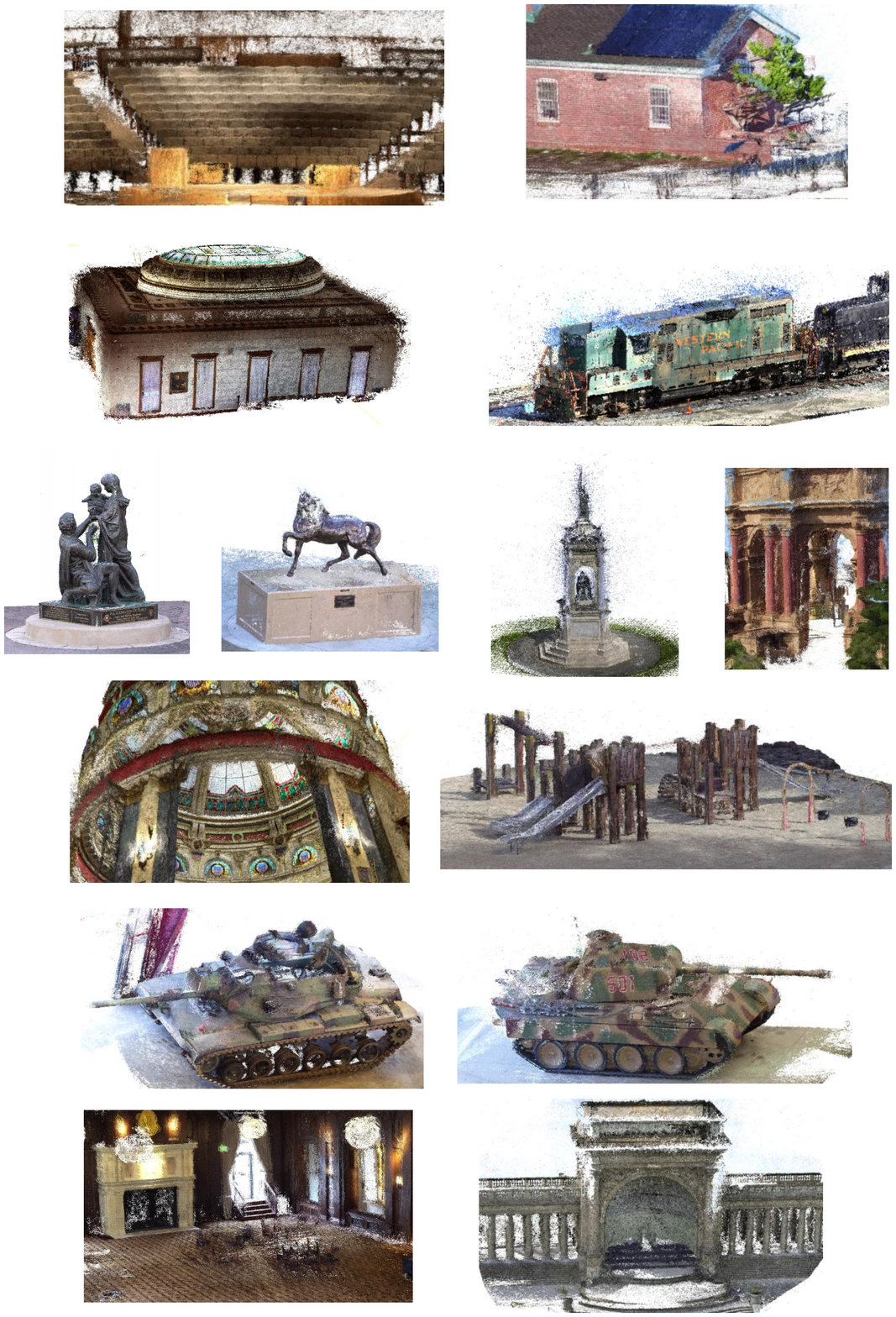}
\caption{Point clouds on Tanks\&Temples~\cite{DBLP:journals/tog/KnapitschPZK17} reconstructed by MVSTER.}
\label{fig:tanks}
\end{figure}

We visualize point cloud reconstruction results of DTU~\cite{DBLP:journals/ijcv/AanaesJVTD16}, ETH3D~\cite{DBLP:conf/cvpr/SchopsSGSSPG17} and Tanks\&Temples~\cite{DBLP:journals/tog/KnapitschPZK17} in Fig. \ref{fig:dtu}, Fig. \ref{fig:eth} and Fig. \ref{fig:tanks}, respectively. MVSTER shows its robustness on scenes with varying input image resolutions and depth ranges.

\par\vfill\par
\clearpage
%
%
\bibliographystyle{splncs04}
\bibliography{submission}

\begin{thebibliography}{10}
\providecommand{\url}[1]{\texttt{#1}}
\providecommand{\urlprefix}{URL }
\providecommand{\doi}[1]{https://doi.org/#1}

\bibitem{DBLP:journals/ijcv/AanaesJVTD16}
Aan{\ae}s, H., Jensen, R.R., Vogiatzis, G., Tola, E., Dahl, A.B.: Large-scale
  data for multiple-view stereopsis. International Journal of Computer Vision
  (2016)

\bibitem{DBLP:conf/acl/AbnarZ20}
Abnar, S., Zuidema, W.H.: Quantifying attention flow in transformers. In:
  Association for Computational Linguistics (2020)

\bibitem{DBLP:journals/corr/ArjovskyCB17}
Arjovsky, M., Chintala, S., Bottou, L.: Wasserstein {GAN}. arXiv preprint
  arXiv:1701.07875  (2017)

\bibitem{bozic2021transformerfusion}
Bozic, A., Palafox, P., Thies, J., Dai, A., Nie{\ss}ner, M.: Transformerfusion:
  Monocular rgb scene reconstruction using transformers. In: Advances in Neural
  Information Processing Systems (2021)

\bibitem{NeillDFCampbell2008UsingMH}
Campbell, N.D.F., Vogiatzis, G., Hern{\'a}ndez, C., Cipolla, R.: Using multiple
  hypotheses to improve depth-maps for multi-view stereo. In: European
  Conference on Computer Vision (2008)

\bibitem{DBLP:conf/eccv/CarionMSUKZ20}
Carion, N., Massa, F., Synnaeve, G., Usunier, N., Kirillov, A., Zagoruyko, S.:
  End-to-end object detection with transformers. In: European Conference on
  Computer Vision (2020)

\bibitem{DBLP:journals/corr/ChenPSA17}
Chen, L., Papandreou, G., Schroff, F., Adam, H.: Rethinking atrous convolution
  for semantic image segmentation. arXiv preprint arXiv:1706.05587  (2017)

\bibitem{DBLP:conf/icml/ChenRC0JLS20}
Chen, M., Radford, A., Child, R., Wu, J., Jun, H., Luan, D., Sutskever, I.:
  Generative pretraining from pixels. In: International Conference on Machine
  Learning (2020)

\bibitem{DBLP:conf/iccv/ChenHXS19}
Chen, R., Han, S., Xu, J., Su, H.: Point-based multi-view stereo network. In:
  IEEE International Conference on Computer Vision (2019)

\bibitem{DBLP:conf/cvpr/ChengXZLLRS20}
Cheng, S., Xu, Z., Zhu, S., Li, Z., Li, L.E., Ramamoorthi, R., Su, H.: Deep
  stereo using adaptive thin volume representation with uncertainty awareness.
  In: IEEE Conference on Computer Vision and Pattern Recognition (2020)

\bibitem{Chitta2021ICCV}
Chitta, K., Prakash, A., Geiger, A.: Neat: Neural attention fields for
  end-to-end autonomous driving. In: IEEE International Conference on Computer
  Vision (2021)

\bibitem{RobertTCollins1996ASA}
Collins, R.T.: A space-sweep approach to true multi-image matching. In: IEEE
  Conference on Computer Vision and Pattern Recognition (1996)

\bibitem{DBLP:conf/nips/Cuturi13}
Cuturi, M.: Sinkhorn distances: Lightspeed computation of optimal transport.
  In: Advances in Neural Information Processing Systems (2013)

\bibitem{DBLP:conf/iccv/DaiQXLZHW17}
Dai, J., Qi, H., Xiong, Y., Li, Y., Zhang, G., Hu, H., Wei, Y.: Deformable
  convolutional networks. In: IEEE International Conference on Computer Vision
  (2017)

\bibitem{DBLP:conf/naacl/DevlinCLT19}
Devlin, J., Chang, M., Lee, K., Toutanova, K.: {BERT:} pre-training of deep
  bidirectional transformers for language understanding. In: Conference of the
  North American Chapter of the Association for Computational Linguistics:
  Human Language Technologies (2019)

\bibitem{DBLP:journals/corr/abs-2111-14600}
Ding, Y., Yuan, W., Zhu, Q., Zhang, H., Liu, X., Wang, Y., Liu, X.:
  Transmvsnet: Global context-aware multi-view stereo network with
  transformers. arXiv preprint arXiv:2111.14600  (2021)

\bibitem{DBLP:conf/iclr/DosovitskiyB0WZ21}
Dosovitskiy, A., Beyer, L., Kolesnikov, A., Weissenborn, D., Zhai, X.,
  Unterthiner, T., Dehghani, M., Minderer, M., Heigold, G., Gelly, S.,
  Uszkoreit, J., Houlsby, N.: An image is worth 16x16 words: Transformers for
  image recognition at scale. In: International Conference on Learning
  Representations (2021)

\bibitem{DBLP:conf/iccv/DosovitskiyFIHH15}
Dosovitskiy, A., Fischer, P., Ilg, E., H{\"{a}}usser, P., Hazirbas, C., Golkov,
  V., van~der Smagt, P., Cremers, D., Brox, T.: Flownet: Learning optical flow
  with convolutional networks. In: IEEE International Conference on Computer
  Vision (2015)

\bibitem{DBLP:conf/iccv/DuggalWMHU19}
Duggal, S., Wang, S., Ma, W., Hu, R., Urtasun, R.: Deeppruner: Learning
  efficient stereo matching via differentiable patchmatch. In: IEEE
  International Conference on Computer Vision (2019)

\bibitem{YasutakaFurukawa2010AccurateDA}
Furukawa, Y., Ponce, J.: Accurate, dense, and robust multiview stereopsis. IEEE
  Transactions on Pattern Analysis and Machine Intelligence  (2010)

\bibitem{DBLP:conf/iccv/GallianiLS15}
Galliani, S., Lasinger, K., Schindler, K.: Massively parallel multiview
  stereopsis by surface normal diffusion. In: IEEE International Conference on
  Computer Vision (2015)

\bibitem{giang2021curvature}
Giang, K.T., Song, S., Jo, S.: Curvature-guided dynamic scale networks for
  multi-view stereo. arXiv preprint arXiv:2112.05999  (2021)

\bibitem{ClmentGodard2017UnsupervisedMD}
Godard, C., Aodha, O.M., Brostow, G.J.: Unsupervised monocular depth estimation
  with left-right consistency. In: IEEE Conference on Computer Vision and
  Pattern Recognition (2017)

\bibitem{DBLP:conf/iccv/GodardAFB19}
Godard, C., Aodha, O.M., Firman, M., Brostow, G.J.: Digging into
  self-supervised monocular depth estimation. In: IEEE International Conference
  on Computer Vision (2019)

\bibitem{DBLP:conf/cvpr/GuFZDTT20}
Gu, X., Fan, Z., Zhu, S., Dai, Z., Tan, F., Tan, P.: Cascade cost volume for
  high-resolution multi-view stereo and stereo matching. In: IEEE Conference on
  Computer Vision and Pattern Recognition (2020)

\bibitem{DBLP:conf/cvpr/HeZH0Z20}
He, C., Zeng, H., Huang, J., Hua, X., Zhang, L.: Structure aware single-stage
  3d object detection from point cloud. In: IEEE Conference on Computer Vision
  and Pattern Recognition (2020)

\bibitem{DBLP:conf/cvpr/HeYFY20}
He, Y., Yan, R., Fragkiadaki, K., Yu, S.: Epipolar transformer for multi-view
  human pose estimation. In: IEEE Conference on Computer Vision and Pattern
  Recognition (2020)

\bibitem{DBLP:conf/cvpr/KeBASB17}
Ke, Q., Bennamoun, M., An, S., Sohel, F.A., Boussa{\"{\i}}d, F.: A new
  representation of skeleton sequences for 3d action recognition. In: IEEE
  Conference on Computer Vision and Pattern Recognition (2017)

\bibitem{DBLP:journals/corr/KingmaB14}
Kingma, D.P., Ba, J.: Adam: {A} method for stochastic optimization. In:
  International Conference on Learning Representations (2015)

\bibitem{DBLP:journals/tog/KnapitschPZK17}
Knapitsch, A., Park, J., Zhou, Q., Koltun, V.: Tanks and temples: benchmarking
  large-scale scene reconstruction. ACM Transactions on Graphics  (2017)

\bibitem{lee2021patchmatchrl}
Lee, J.Y., DeGol, J., Zou, C., Hoiem, D.: Patchmatch-rl: Deep mvs with
  pixelwise depth, normal, and visibility. In: IEEE International Conference on
  Computer Vision (2021)

\bibitem{li2021revisiting}
Li, Z., Liu, X., Drenkow, N., Ding, A., Creighton, F.X., Taylor, R.H.,
  Unberath, M.: Revisiting stereo depth estimation from a sequence-to-sequence
  perspective with transformers. In: IEEE International Conference on Computer
  Vision (2021)

\bibitem{DBLP:conf/cvpr/LinDGHHB17}
Lin, T., Doll{\'{a}}r, P., Girshick, R.B., He, K., Hariharan, B., Belongie,
  S.J.: Feature pyramid networks for object detection. In: IEEE Conference on
  Computer Vision and Pattern Recognition (2017)

\bibitem{liu2021Swin}
Liu, Z., Lin, Y., Cao, Y., Hu, H., Wei, Y., Zhang, Z., Lin, S., Guo, B.: Swin
  transformer: Hierarchical vision transformer using shifted windows. IEEE
  International Conference on Computer Vision  (2021)

\bibitem{DBLP:conf/cvpr/LuoH21}
Luo, S., Hu, W.: Diffusion probabilistic models for 3d point cloud generation.
  In: IEEE Conference on Computer Vision and Pattern Recognition (2021)

\bibitem{ma2021epp}
Ma, X., Gong, Y., Wang, Q., Huang, J., Chen, L., Yu, F.: Epp-mvsnet:
  Epipolar-assembling based depth prediction for multi-view stereo. In: IEEE
  International Conference on Computer Vision (2021)

\bibitem{DBLP:conf/eccv/MildenhallSTBRN20}
Mildenhall, B., Srinivasan, P.P., Tancik, M., Barron, J.T., Ramamoorthi, R.,
  Ng, R.: Nerf: Representing scenes as neural radiance fields for view
  synthesis. In: European Conference on Computer Vision (2020)

\bibitem{DBLP:conf/nips/MordanTHC18}
Mordan, T., Thome, N., H{\'{e}}naff, G., Cord, M.: Revisiting multi-task
  learning with {ROCK:} a deep residual auxiliary block for visual detection.
  In: Advances in Neural Information Processing Systems (2018)

\bibitem{DBLP:journals/corr/abs-2201-01501}
Peng, R., Wang, R., Wang, Z., Lai, Y., Wang, R.: Rethinking depth estimation
  for multi-view stereo: {A} unified representation and focal loss. arXiv
  preprint arXiv:2201.01501  (2022)

\bibitem{DBLP:journals/ftml/PeyreC19}
Peyr{\'{e}}, G., Cuturi, M.: Computational optimal transport. Foundations and
  Trends in Machine Learning  (2019)

\bibitem{DBLP:conf/cvpr/QiSMG17}
Qi, C.R., Su, H., Mo, K., Guibas, L.J.: Pointnet: Deep learning on point sets
  for 3d classification and segmentation. In: IEEE Conference on Computer
  Vision and Pattern Recognition (2017)

\bibitem{DBLP:conf/nips/QiYSG17}
Qi, C.R., Yi, L., Su, H., Guibas, L.J.: Pointnet++: Deep hierarchical feature
  learning on point sets in a metric space. In: Advances in Neural Information
  Processing Systems (2017)

\bibitem{radford2018improving}
Radford, A., Narasimhan, K., Salimans, T., Sutskever, I.: Improving language
  understanding by generative pre-training. OpenAI Preprint  (2018)

\bibitem{DBLP:conf/miccai/RonnebergerFB15}
Ronneberger, O., Fischer, P., Brox, T.: U-net: Convolutional networks for
  biomedical image segmentation. In: Medical Image Computing and
  Computer-Assisted Intervention (2015)

\bibitem{DBLP:conf/cvpr/SchonbergerF16}
Sch{\"{o}}nberger, J.L., Frahm, J.: Structure-from-motion revisited. In: IEEE
  Conference on Computer Vision and Pattern Recognition (2016)

\bibitem{DBLP:conf/cvpr/SchopsSGSSPG17}
Sch{\"{o}}ps, T., Sch{\"{o}}nberger, J.L., Galliani, S., Sattler, T.,
  Schindler, K., Pollefeys, M., Geiger, A.: A multi-view stereo benchmark with
  high-resolution images and multi-camera videos. In: IEEE Conference on
  Computer Vision and Pattern Recognition (2017)

\bibitem{DBLP:conf/cvpr/ShenDR21}
Shen, Z., Dai, Y., Rao, Z.: Cfnet: Cascade and fused cost volume for robust
  stereo matching. In: IEEE Conference on Computer Vision and Pattern
  Recognition (2021)

\bibitem{DBLP:conf/cvpr/ShiGJ0SWL20}
Shi, S., Guo, C., Jiang, L., Wang, Z., Shi, J., Wang, X., Li, H.: {PV-RCNN:}
  point-voxel feature set abstraction for 3d object detection. In: IEEE
  Conference on Computer Vision and Pattern Recognition (2020)

\bibitem{sinha2020deltas}
Sinha, A., Murez, Z., Bartolozzi, J., Badrinarayanan, V., Rabinovich, A.:
  Deltas: Depth estimation by learning triangulation and densification of
  sparse points. In: European Conference on Computer Vision (2020)

\bibitem{DBLP:conf/cvpr/TankovichH0KFB21}
Tankovich, V., Hane, C., Zhang, Y., Kowdle, A., Fanello, S.R., Bouaziz, S.:
  Hitnet: Hierarchical iterative tile refinement network for real-time stereo
  matching. In: IEEE Conference on Computer Vision and Pattern Recognition. pp.
  14362--14372 (2021)

\bibitem{DBLP:conf/acl/TenneyDP19}
Tenney, I., Das, D., Pavlick, E.: {BERT} rediscovers the classical {NLP}
  pipeline. In: Association for Computational Linguistics (2019)

\bibitem{DBLP:journals/mva/TolaSF12}
Tola, E., Strecha, C., Fua, P.: Efficient large-scale multi-view stereo for
  ultra high-resolution image sets. Machine Vision and Applications  (2012)

\bibitem{DBLP:conf/nips/VaswaniSPUJGKP17}
Vaswani, A., Shazeer, N., Parmar, N., Uszkoreit, J., Jones, L., Gomez, A.N.,
  Kaiser, L., Polosukhin, I.: Attention is all you need. In: Advances in Neural
  Information Processing Systems (2017)

\bibitem{DBLP:journals/corr/abs-2112-05126}
Wang, F., Galliani, S., Vogel, C., Pollefeys, M.: Itermvs: Iterative
  probability estimation for efficient multi-view stereo. arXiv preprint
  arXiv:2112.05126  (2021)

\bibitem{DBLP:conf/cvpr/WangGVSP21}
Wang, F., Galliani, S., Vogel, C., Speciale, P., Pollefeys, M.: Patchmatchnet:
  Learned multi-view patchmatch stereo. In: IEEE Conference on Computer Vision
  and Pattern Recognition (2021)

\bibitem{DBLP:conf/eccv/WangZGAYC20}
Wang, H., Zhu, Y., Green, B., Adam, H., Yuille, A.L., Chen, L.: Axial-deeplab:
  Stand-alone axial-attention for panoptic segmentation. In: European
  Conference on Computer Vision (2020)

\bibitem{DBLP:conf/cvpr/WatsonAPBF21}
Watson, J., Aodha, O.M., Prisacariu, V., Brostow, G.J., Firman, M.: The
  temporal opportunist: Self-supervised multi-frame monocular depth. In: IEEE
  Conference on Computer Vision and Pattern Recognition (2021)

\bibitem{wei2021aa}
Wei, Z., Zhu, Q., Min, C., Chen, Y., Wang, G.: Aa-rmvsnet: Adaptive aggregation
  recurrent multi-view stereo network. In: IEEE International Conference on
  Computer Vision (2021)

\bibitem{DBLP:conf/cvpr/XuT19}
Xu, Q., Tao, W.: Multi-scale geometric consistency guided multi-view stereo.
  In: IEEE Conference on Computer Vision and Pattern Recognition (2019)

\bibitem{DBLP:conf/aaai/XuT20}
Xu, Q., Tao, W.: Learning inverse depth regression for multi-view stereo with
  correlation cost volume. In: AAAI Conference on Artificial Intelligence
  (2020)

\bibitem{DBLP:journals/corr/abs-2007-07714}
Xu, Q., Tao, W.: Pvsnet: Pixelwise visibility-aware multi-view stereo network.
  arXiv preprint arXiv:2007.07714  (2020)

\bibitem{DBLP:conf/eccv/YanWYDZCWT20}
Yan, J., Wei, Z., Yi, H., Ding, M., Zhang, R., Chen, Y., Wang, G., Tai, Y.:
  Dense hybrid recurrent multi-view stereo net with dynamic consistency
  checking. In: European Conference on Computer Vision (2020)

\bibitem{DBLP:conf/cvpr/YangYFLG20}
Yang, F., Yang, H., Fu, J., Lu, H., Guo, B.: Learning texture transformer
  network for image super-resolution. In: IEEE Conference on Computer Vision
  and Pattern Recognition (2020)

\bibitem{DBLP:conf/cvpr/YangMAL20}
Yang, J., Mao, W., Alvarez, J.M., Liu, M.: Cost volume pyramid based depth
  inference for multi-view stereo. In: IEEE Conference on Computer Vision and
  Pattern Recognition (2020)

\bibitem{DBLP:conf/cvpr/YangLLYMHP21}
Yang, W., Li, Q., Liu, W., Yu, Y., Ma, Y., He, S., Pan, J.: Projecting your
  view attentively: Monocular road scene layout estimation via cross-view
  transformation. In: IEEE Conference on Computer Vision and Pattern
  Recognition (2021)

\bibitem{DBLP:conf/eccv/YaoLLFQ18}
Yao, Y., Luo, Z., Li, S., Fang, T., Quan, L.: Mvsnet: Depth inference for
  unstructured multi-view stereo. In: European Conference on Computer Vision
  (2018)

\bibitem{DBLP:conf/cvpr/0008LLSFQ19}
Yao, Y., Luo, Z., Li, S., Shen, T., Fang, T., Quan, L.: Recurrent mvsnet for
  high-resolution multi-view stereo depth inference. In: IEEE Conference on
  Computer Vision and Pattern Recognition (2019)

\bibitem{DBLP:conf/cvpr/0008LLZRZFQ20}
Yao, Y., Luo, Z., Li, S., Zhang, J., Ren, Y., Zhou, L., Fang, T., Quan, L.:
  Blendedmvs: {A} large-scale dataset for generalized multi-view stereo
  networks. In: IEEE Conference on Computer Vision and Pattern Recognition
  (2020)

\bibitem{DBLP:conf/eccv/YiWDZCWT20}
Yi, H., Wei, Z., Ding, M., Zhang, R., Chen, Y., Wang, G., Tai, Y.: Pyramid
  multi-view stereo net with self-adaptive view aggregation. In: European
  Conference on Computer Vision (2020)

\bibitem{DBLP:conf/cvpr/YuG20}
Yu, Z., Gao, S.: Fast-mvsnet: Sparse-to-dense multi-view stereo with learned
  propagation and gauss-newton refinement. In: IEEE Conference on Computer
  Vision and Pattern Recognition (2020)

\bibitem{DBLP:conf/bmvc/ZhangYLLF20}
Zhang, J., Yao, Y., Li, S., Luo, Z., Fang, T.: Visibility-aware multi-view
  stereo network. In: British Machine Vision Conference (2020)

\bibitem{DBLP:conf/wacv/ZhangHW0021}
Zhang, X., Hu, Y., Wang, H., Cao, X., Zhang, B.: Long-range attention network
  for multi-view stereo. In: {IEEE} Winter Conference on Applications of
  Computer Vision (2021)

\bibitem{DBLP:conf/cvpr/ZhaoZZZ19}
Zhao, M., Zhang, J., Zhang, C., Zhang, W.: Leveraging heterogeneous auxiliary
  tasks to assist crowd counting. In: IEEE Conference on Computer Vision and
  Pattern Recognition (2019)

\bibitem{DBLP:conf/cvpr/ZhouT18}
Zhou, Y., Tuzel, O.: Voxelnet: End-to-end learning for point cloud based 3d
  object detection. In: IEEE Conference on Computer Vision and Pattern
  Recognition (2018)

\bibitem{DBLP:journals/corr/abs-2112-00336}
Zhu, J., Peng, B., Li, W., Shen, H., Zhang, Z., Lei, J.: Multi-view stereo with
  transformer. arXiv preprint arXiv:2112.00336  (2021)

\end{thebibliography}
\end{document}